\documentclass{article} % For LaTeX2e
\usepackage{iclr2026_conference,times}

\usepackage{amsmath,amsfonts,bm}

\def\eqref#1{equation~\ref{#1}}
\def\1{\bm{1}}

\DeclareMathAlphabet{\mathsfit}{\encodingdefault}{\sfdefault}{m}{sl}
\SetMathAlphabet{\mathsfit}{bold}{\encodingdefault}{\sfdefault}{bx}{n}

\usepackage{hyperref}
\usepackage{url}
\usepackage{graphicx}
\usepackage{booktabs}
\usepackage{multirow}
\usepackage{wrapfig}
\usepackage[table]{xcolor}
\definecolor{lightblue}{RGB}{230, 235, 250}
\definecolor{lightgray}{gray}{0.9}

\title{A State-Transition Framework for Efficient LLM Reasoning}

\author{
Liang Zhang$^{1,3}$, Yu Zhao$^{2}$, Longyue Wang$^{2}$, Tianqi Shi$^{2}$, Weihua Luo$^{2}$,\\
\bfseries{ Kaifu Zhang$^{2}$, Jinsong Su$^{1,3,4}$\thanks{Corresponding author.}} \\
$^{1}$School of Informatics, Xiamen University, China~ 
$^{2}$Alibaba International Digital Commerce \\
$^{3}$Key Laboratory of Digital Protection and Intelligent Processing of Intangible Cultural\\ ~~Heritage of Fujian and Taiwan (Xiamen University), Ministry of Culture and Tourism, China\\
$^{4}$Shanghai Artificial Intelligence Laboratory \\
\texttt{lzhang@stu.xmu.edu.cn, jssu@xmu.edu.cn}\\
\texttt{\{fengli.zy, wanglongyue.wly, weihua.luowh\}@alibaba-inc.com}
}

\iclrfinalcopy % Uncomment for camera-ready version, but NOT for submission.
\begin{document}

\maketitle

\begin{abstract}
While Long Chain-of-Thought (CoT) reasoning significantly improves Large Language Models (LLMs) performance on complex reasoning tasks, the substantial computational and memory costs of generating long CoT sequences limit their efficiency and practicality.
Existing studies usually enhance the reasoning efficiency of LLMs by compressing CoT sequences.
However, this approach conflicts with test‑time scaling, limiting the reasoning capacity of LLMs.
In this paper, we propose an efficient reasoning framework that models the reasoning process of LLMs as a \textit{state‑transition} process.
Specifically, we first apply a linear attention mechanism to estimate the LLM’s reasoning state, which records the historical reasoning information from previous reasoning steps.
Then, based on the query prompt and the reasoning state, the LLM can efficiently perform the current reasoning step and update the state.
With the linear attention, each token in the current reasoning step can directly retrieve relevant historical reasoning information from the reasoning state, without explicitly attending to tokens in previous reasoning steps.
In this way, the computational complexity of attention is reduced from quadratic to linear, significantly improving the reasoning efficiency of LLMs.
In addition, we propose a state-based reasoning strategy to mitigate the over-thinking issue caused by noisy reasoning steps.
Extensive experiments across multiple datasets and model sizes demonstrate that our framework not only improves the reasoning efficiency of LLMs but also enhances their reasoning performance.

\end{abstract}

\section{introduction}
Chain‑of‑Thought (CoT) \citep{c:1,c:2} has become a core technique for enhancing the reasoning ability of large language models (LLMs) on complex tasks.
% such as mathematical problem solving and code generation.
Through prompting step-by-step reasoning, CoT enables LLMs to decompose complex problems into simpler subtasks, thus improving their problem-solving capabilities \citep{c:3,c:4,c:5,c:79}.
% Chain‑of‑Thought (CoT) \citep{c:1,c:2} has become a core technique for enhancing the reasoning ability of large language models (LLMs), because by prompting step‑by‑step reasoning CoT enables LLMs to decompose complex problems into simpler subtasks and thereby improve their problem‑solving capabilities \citep{c:3,c:4,c:5}.
Recent studies, including OpenAI o1 \citep{c:6},  QwQ \citep{c:8}, and DeepSeek‑R1 \citep{c:7}, demonstrate that scaling up CoT length can further enhance the reasoning abilities of LLMs.
However, since most current LLMs are built on the Transformer architecture \citep{c:9}, the computational complexity of their attention grows quadratically with context length, and the memory overhead of their KV‑cache increases linearly with context length. 
% For example, when the context length reaches $10^4$, the KV‑cache of Qwen2.5‑32B occupies memory comparable to the size of the model itself \citep{c:10}.
Hence, generating long CoT substantially increase the computational and memory cost of LLMs, limiting their practical efficiency on complex reasoning tasks.

To improve the reasoning efficiency of LLMs, previous studies employ prompting \citep{c:12,c:14}, supervised fine‑tuning (SFT) \citep{c:15,c:25}, or reinforcement learning (RL) \citep{c:26,c:27} to encourage LLMs toward generating shorter CoT sequences.
However, these methods often impair the reasoning ability of LLMs \citep{c:16,c:17}, since CoT shortening conflicts with test‑time scaling \citep{c:6}.
To preserve the reasoning ability of LLMs, some studies \citep{c:23,c:22,c:21}  express the CoT in more concise text (\textit{e.g.}, by removing less important tokens or rewriting with GPT‑4) to reduce its length.
However, they risk losing critical reasoning information or reducing interpretability when simplifying long CoT \citep{c:24}.

% To alleviate this challenge, existing methods for CoT compression can be roughly divided into three categories. 
% \textbf{The first category} leverages prompting \citep{c:12,c:14}, supervised fine‑tuning (SFT) \citep{c:15,c:25}, and reinforcement learning (RL) \citep{c:26,c:27} to encourage LLMs to generate shorter CoT sequences.
% Since such CoT reductions conflict with test-time scaling \citep{c:6}, these methods often impair the reasoning ability of LLMs \citep{c:16,c:17}.
% \textbf{The second category} compresses information from multiple text tokens into a single hidden vector, converting longer discrete CoT sequences into shorter continuous ones to improve the reasoning efficiency of LLMs \citep{c:20,c:19}.
% However, such methods typically require altering the input/output patterns of LLMs, increasing their vulnerability to catastrophic forgetting and limiting their performance.
% \textbf{The third category} expresses the CoT in more concise text (\textit{e.g.}, by removing less important tokens or rewriting with GPT‑4) to reduce its length while preserving the reasoning ability of LLMs \citep{c:23,c:22,c:21}.
% Nevertheless, when simplifying long CoT, these methods risk losing critical reasoning information or reducing interpretability \citep{c:24}.

In this paper, we propose an efficient reasoning framework for LLMs, which models the reasoning process of LLMs as a state‑transition process.
We regard a long CoT as a sequence of reasoning steps, where LLMs perform a specific thinking pattern in each step, such as \textit{induction} or \textit{reflection}.
Notably, each reasoning step contains two types of information: substantial linguistic information to ensure its fluency, and limited reasoning information to support subsequent reasoning or answer generation \citep{c:10,c:21}.
% Notably, each reasoning step predominantly contains linguistic information to ensure its fluency, and a smaller portion of reasoning information to support subsequent reasoning or answer generation \citep{c:10,c:21}.
Thus, our framework (Figure~\ref{fig1}) first compresses the reasoning information from previously generated reasoning steps into a matrix, termed as the \textit{reasoning state matrix}.
Then, based on the query prompt and the state matrix, LLMs can efficiently generate the current reasoning step and updates the state matrix accordingly.
% With the reasoning state, tokens in the current reasoning step bypass explicit attention to tokens from earlier steps, reducing the computation complexity of attention from quadratic to linear and improving reasoning efficiency.
Specifically, tokens in the current reasoning step can directly retrieve relevant historical reasoning information from the reasoning state, without explicitly attending to tokens in previous steps.
In this way, we effectively reduce both the computational complexity of attention and the memory overhead of the KV‑cache.
% Moreover, this also avoid caching the KV vectors of tokens from completed reasoning steps, which greatly reduces memory consumption.
% Meanwhile, in our framework, each token attends only to the tokens in the query prompt and ones in its current reasoning step, reducing attention computation complexity from quadratic to linear and improving reasoning efficiency.
Crucially, our framework does not shorten or simplify the CoT sequences generated by LLMs, thus preserving their reasoning ability and interpretability.

To efficiently obtain the state matrix, we design a Mixed Attention Module (\textbf{MAM}) to replace the original attention module in LLMs, which consists of a Softmax‑Attention (\textbf{SA}) submodule and a Linear‑Attention (\textbf{LA}) submodule.
We use the original attention module of LLMs as our SA submodule, where each token can only attend to the tokens in the query prompt and those in its current reasoning step.
In the LA submodule, we adopt a linear‑attention mechanism \citep{c:28} to capture the reasoning state of LLMs.
Meanwhile, with the linear attention, each token can directly retrieve relevant historical reasoning information from the reasoning state.
Compared with other methods, such as CNN \citep{c:29} and Q‑Former \citep{c:30}, linear attention offers the following advantages in capturing the model’s reasoning state:
(1) As a variant of softmax attention, linear attention is naturally compatible with it, thereby reducing the risk of losing critical reasoning information during compression and ensuring the stability and efficiency of model training.
(2) Recent studies \citep{c:31,c:32} have demonstrated that the state-update process of linear attention is essentially a gradient‑descent learning procedure (i.e., test‑time training), revealing its strong potential for handling complex reasoning tasks \citep{c:42}.

% Recent studies \citep{c:33,c:34} indicate 
Another challenge is that LLMs often produce noisy reasoning steps, which may mislead subsequent reasoning steps and lead to the overthinking issue \citep{c:33,c:34}.
To mitigate this issue, we propose a \textit{state‑based reasoning strategy}.
Given the model’s state‑transition process is a gradient‑descent process, we first apply the momentum method to accumulate gradients from completed reasoning steps, obtaining a global gradient.
The global gradient indicates the global reasoning direction of LLMs.
Thus, we employ it to guide LLMs in completing the current reasoning step, ensuring they do not significantly deviate from the global direction.
% By doing so, we not only mitigate the adverse effects of noisy reasoning steps on the model’s reasoning process, but also preserve its capacity to explore during inference.

To validate the efficacy of our framework, we conduct experiments on {\em seven widely-used benchmark datasets}.
Experimental results show that our framework not only improves the reasoning efficiency of LLMs but also enhances their reasoning performance.
Extensive ablation studies further demonstrate the effectiveness of various components in our framework.

\section{Preliminary}
We first present a brief background on linear attention, which lays the foundation for our proposed framework.
Meanwhile, we outline the specific form of the CoT sequences in our framework.

\subsection{Linear Attention}
\textbf{Softmax Attention (SA).}
\label{s:2.1}
% Popular LLMs, such as Qwen 3 \citep{c:36} and Llama 4 \citep{c:35}, are decoder‑only Transformers composed of repeated blocks of multi‑head softmax attention followed by FFN \citep{c:37}.
Popular LLMs, such as Qwen 3 \citep{c:36} and Llama 4 \citep{c:35}, adopt a decoder‑only Transformer architecture composed of repeated blocks of multi‑head softmax attention followed by feed‑forward networks (FFNs) \citep{c:37}.
Given the input sequence $\boldsymbol{X} {=}[x_1, \cdots , x_{|\boldsymbol{X}|}]$,  the softmax attention can be formulated as follows:

\begin{equation}
\begin{aligned}
\boldsymbol{o}_t =
\frac{\sum_{i=1}^t \exp\left( \boldsymbol{q}_t \boldsymbol{k}_i^\top / \sqrt{d} \right) \boldsymbol{v}_i}
{\sum_{i'=1}^t \exp\left( \boldsymbol{q}_t \boldsymbol{k}_{i'}^\top / \sqrt{d} \right)}; \;\;\;\;
 \boldsymbol{q}_t, \boldsymbol{k}_t, \boldsymbol{v}_t = x_t \boldsymbol{W}_Q, \; x_t \boldsymbol{W}_K, \; x_t \boldsymbol{W}_V, \\
\end{aligned}
\label{q:1}
\end{equation}
where $\boldsymbol{W}_Q$, $\boldsymbol{W}_K$, $\boldsymbol{W}_V$ are learnable weight matrices.
The computational complexity of the softmax function increases quadratically with the length of the context sequence.
Moreover, softmax attention heavily depends on the growing KV-cache to recall historical information for sequence modeling, leading to substantial memory overheads, particularly in the long context setting.

\textbf{Linear Attention (LA).} Linear attention is a variant of softmax attention, designed to reduce its computational complexity and memory costs.
Here, we first replaces the exponential function $\exp(\cdot)$ in softmax attention with a simpler kernel function $\phi(\cdot)$: $\exp(\boldsymbol{q}_t \boldsymbol{k}_i/\sqrt{d}) \to \phi(\boldsymbol{q}_t) \phi(\boldsymbol{k}_i)^{\top}$.
Next,  based on the associative property of matrix products, linear attention can be written as
\begin{equation}
\begin{aligned}
&\boldsymbol{o}_t =
\frac{\sum_{i=1}^t \phi(\boldsymbol{q}_t) \phi(\boldsymbol{k}_i)^{\top} \boldsymbol{v}_i}
{\sum_{i'=1}^t \phi(\boldsymbol{q}_t) \phi(\boldsymbol{k}_{i'})^{\top}} = \frac{\phi(\boldsymbol{q}_t) \sum_{i=1}^t \phi(\boldsymbol{k}_i)^{\top} \boldsymbol{v}_i}
{\phi(\boldsymbol{q}_t) \sum_{i'=1}^t  \phi(\boldsymbol{k}_{i'})^{\top}}. 
\end{aligned}
\end{equation}
% Although various kernels have been explored \citep{c:38,c:39}, recent studies \citep{c:40,c:46} have shown that setting $\phi(\cdot)$ to the identity function without a normalizer performs well in practice: 
Moreover, recent studies \citep{c:40,c:46} have shown that linear attention can perform well even when $\phi(\cdot)$ is set to the identity function and the normalizer is removed:
\begin{equation}
\begin{aligned}
\boldsymbol{o}_t = \boldsymbol{q}_t \sum_{i=1}^t \boldsymbol{k}_i^{\top} \boldsymbol{v}_i = \boldsymbol{q}_t \boldsymbol{S}_t; \;\;\;\;  \boldsymbol{S}_t {=} \sum_{i=1}^t \boldsymbol{k}_i^{\top} \boldsymbol{v}_i, \\
\end{aligned}
\label{q:2}
\end{equation}
where $\boldsymbol{S}_t$ is the state matrix storing historical information.
By using $\boldsymbol{S}_t$, linear attention can perform sequence modeling in linear time while maintaining constant memory overhead.
% The state matrix $\boldsymbol{S}_t$ allows linear attention to perform sequence modeling in linear time while maintaining constant memory overhead.
% With the state matrix $\boldsymbol{S}_t$, linear attention achieves efficient linear-time sequence modeling while maintaining constant memory consumption.

\textbf{The Test-Time Training (TTT) Perspective of LA.}
Recent works \citep{c:41,c:33} treat the state matrix $\boldsymbol{S}_t$ in linear attention as a fast-adapting parameter, updated at every token via lightweight gradient descent.
They further introduce an online learning objective for linear attention and provide its closed-form gradient descent solution:
\begin{equation}
\begin{aligned}
&\textbf{Objective:} \;\;\;
\mathcal{L}(\boldsymbol{S}) = - \langle \boldsymbol{S} \boldsymbol{k}_t, \boldsymbol{v}_t\rangle, \\
&\textbf{SGD update:}\;\;\;
\boldsymbol{S}_t = \boldsymbol{S}_{t-1} - \beta \nabla \mathcal{L}_t(\boldsymbol{S}_{t-1})=\boldsymbol{S}_{t-1} + \beta \boldsymbol{v}_t \boldsymbol{k}_{i}^\top, \\
\end{aligned}
\label{q:3}
\end{equation}
where $\langle \cdot,\cdot\rangle$ denotes the inner product, and $\beta$ is the learning rate.
When $\beta$ is set to $1$, the state update process of linear attention is equivalent to training the state matrix $\boldsymbol{S}_{t}$ using the learning objective $\mathcal{L}(\boldsymbol{S})$.
While current studies have yet to provide direct evidence that linear attention can improve the model’s reasoning ability, its theoretical properties suggest significant potential \citep{c:42}.

\begin{figure}[t]
\setlength{\belowcaptionskip}{-8pt}
\centering
\includegraphics[width=0.9 \textwidth]{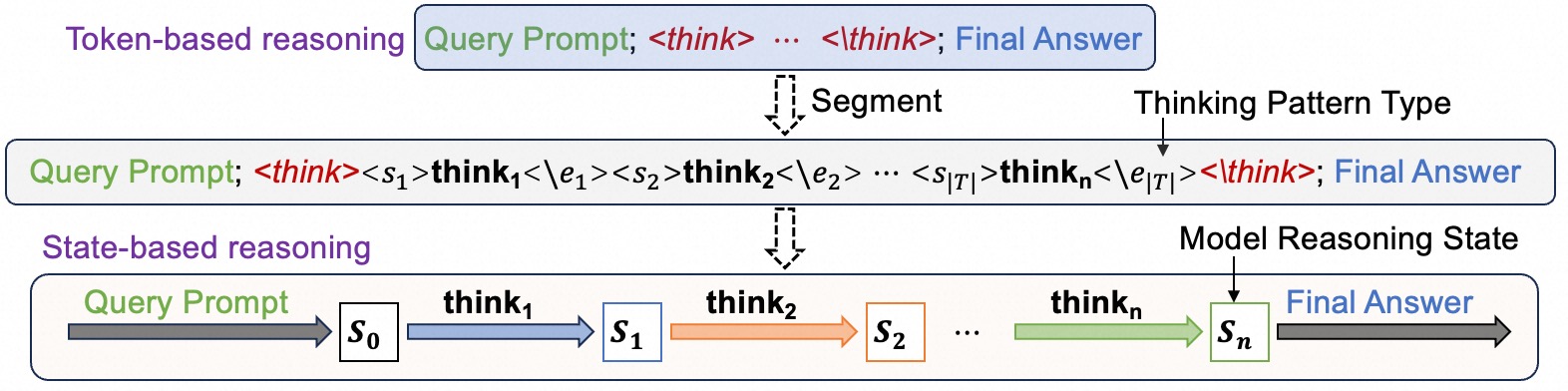}
\caption{A comparison of traditional {\em token‑based reasoning} with our {\em state‑based reasoning}. $\textbf{think}_{t}$ denotes one reasoning step. 
In state-based reasoning, LLMs can efficiently generate $\textbf{think}_{t}$ only based on the query prompt and the reasoning state $\boldsymbol{S}_{t{-}1}$.
% During the state-based reasoning phase, each token in $\textbf{think}_{t+1}$ requires only \emph{query prompt} and \emph{current reasoning state} $\boldsymbol{S}_{t}$.
}
\label{fig1}
\end{figure}

\subsection{Long CoT Segmentation}
% During reasoning, LLMs engage in various thinking patterns, such as Reflection and Result Verification, and often employ common transitional tokens (\textit{e.g.}, “Wait” and “Hmm”) to switch between them \citep{c:34,c:43}.
A long CoT sample $(\boldsymbol{Q},\boldsymbol{T},\boldsymbol{A})$ usually comprises three parts: a query prompt $\boldsymbol{Q}$,  a thinking sequence $\boldsymbol{T}$, and a final answer $\boldsymbol{A}$.
Given the query prompt $\boldsymbol{Q}$, LLMs first perform complex reasoning ($\boldsymbol{T}$), and then generates the final answer $\boldsymbol{A}$.
During reasoning, LLMs engage in various thinking patterns (\textit{e.g.}, \textit{reflection} and \textit{result verification}) and switch between them using common transitional tokens (\textit{e.g.}, “\textit{Wait}”, “\textit{Hmm}”) \citep{c:34,c:43}.
Recent work \citep{c:44} further shows that LLMs usually transition thinking patterns at some high‑entropy tokens, such as “\textit{Alternatively}” and “\textit{Maybe}”.
Here, we refer to the completion of a thinking pattern by LLMs as a \textit{reasoning step}.
Notably, when performing the current reasoning step, LLMs only consider the reasoning information (\textit{e.g.}, conclusion) of previous completed steps without their linguistic information (\textit{e.g.}, grammar).
% This suggests that LLMs tend to perform each reasoning step in a relatively independent manner.
% Therefore, our framework manages the reasoning process of LLMs in units of reasoning steps.

Following \cite{c:44}, we first extract high‑entropy transition tokens from the long CoT training set,  which occur at the start of a sentence.
Next, we use these tokens to segment the thinking sequence $\boldsymbol{T}$ in each training sample into a sequence of reasoning steps.
Meanwhile, we cluster all reasoning steps in the entire training set, with each cluster corresponding to a thinking pattern (\textit{i.e.}, type).
% Meanwhile, we cluster all reasoning steps in the training set to identify the thinking pattern (\textit{i.e.}, type) of each step.
Finally, for each thinking pattern, we introduce two special tokens to enclose its corresponding reasoning steps, as shown in Figure~\ref{fig1}.
With the reconstructed training set, the trained LLMs can generate more structured thinking sequences.
Meanwhile, these special tokens allow us to track and control the reasoning processes of LLMs more accurately.
% A generalization analysis and further details of our CoT segmentation method are provided in Appendix~\ref{appendix:G} and \ref{appendix:H}. 

\section{Our framework}
\begin{figure}[t]
\centering
\includegraphics[width=0.8 \textwidth]{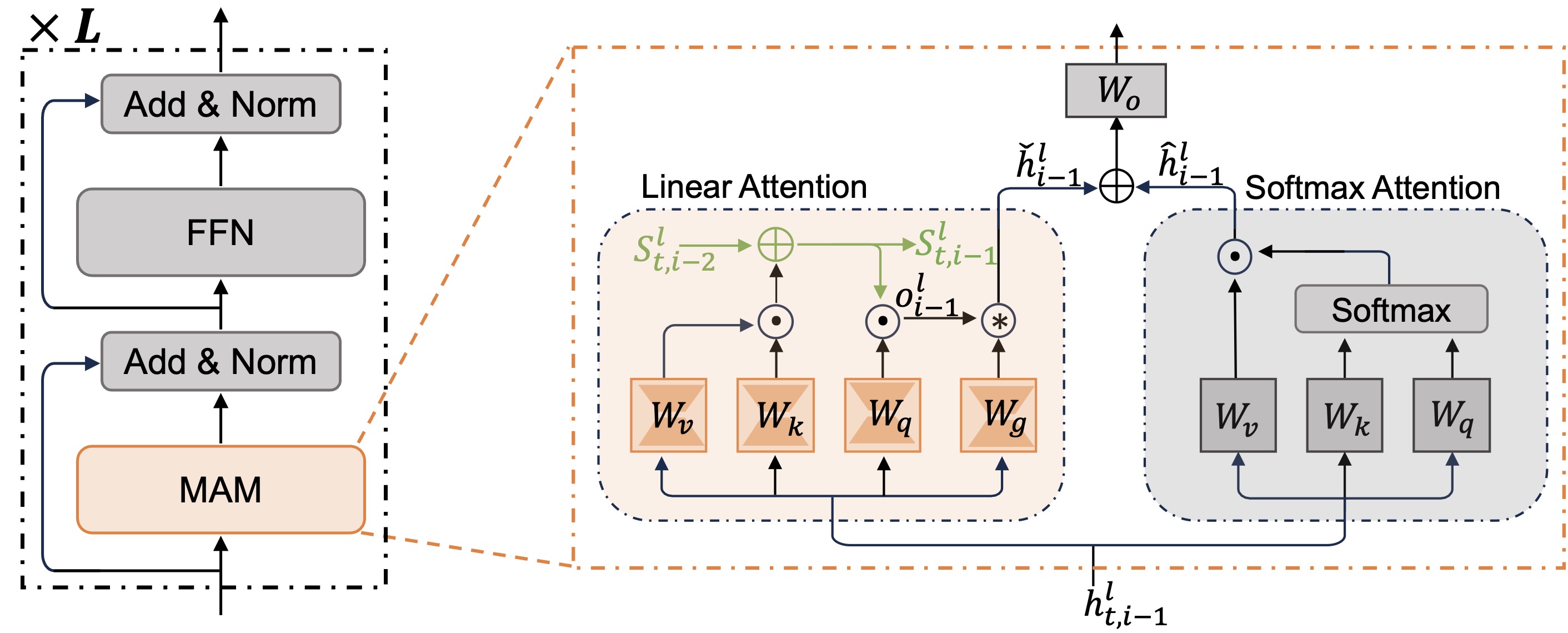}
\caption{In our framework, the original softmax attention module in LLMs is replaced with our mixed attention module (MAM), thus improving reasoning efficiency and reducing memory cost.
}
\label{fig2}
\end{figure}

In this section, we provide a detailed description for our proposed framework.
In our framework, we first design a mixed attention module (\textbf{MAM}) to replace the original attention module in LLMs, as illustrated in Figure~\ref{fig2}.
Using MAM, we can model the reasoning process of LLMs as a state‑transition process (see Section~\ref{S:3.1}).
% Specifically, we first elaborate on  our designed Mixed Attention Module \textbf{MAM} in Section~\ref{S:3.1}.
Then, we further introduces our state‑based reasoning strategy in Section~\ref{S:3.2}.
Finally, our training strategy is presented in Section \ref{S:3.3}.
% Finally, we detail our MAM training strategy in Section~\ref{S:3.3}.

\subsection{ Mixed Attention Module}
\label{S:3.1}

Our framework aims to improve the reasoning efficiency of LLMs without sacrificing reasoning ability by modeling their reasoning process as a state‑transition process.
To achieve this, we design a MAM to \textit{replace} the softmax attention module in LLMs, which consists of a Softmax Attention (\textbf{SA}) submodule and a Linear Attention (\textbf{LA}) submodule.
To avoid the performance loss caused by this replacement, we use the original softmax attention module of LLMs as our SA submodule.
However, in the SA submodule, each token can only attend to the tokens in the query prompt $\boldsymbol{Q}$ and the previously generated tokens in its reasoning step.
By doing so, we reduce the computational complexity of attention from quadratic $O(\mathcal{C}^2)$ to linear $O(\mathcal{C})$ and the memory usage of the KV‑cache from linear $O(\mathcal{C})$ to constant $O(1)$, where $\mathcal{C}$ denotes the context length.
This significantly improves the reasoning efficiency of our model, especially in complex scenarios requiring longer thinking sequences.
Moreover, the LA submodule applies a linear attention mechanism to obtain the LLM’s reasoning state matrix, which records the reasoning information from previously completed reasoning steps.
Therefore, each token in the current reasoning step can access relevant historical information from the state matrix without attending directly to tokens in previous reasoning steps.

Given our LLM generates each token in every reasoning step using the same procedure, we take the generation of the $i$-th token $x_{t,i}$ in the $t$-th reasoning step as an example.
Specifically, at the $l$-th layer of our model, we first feed $h_{t,i-1}^{(l-1)}$, the output feature of the input token $x_{t,i-1}$ at the ($l$-1)-th layer, into the MAM of the current layer.
Then, $h_{t,i-1}^{(l-1)}$ is passed to the two submodules in MAM:

% \textbf{In SA submodule}, the KV vectors of tokens from completed reasoning steps are removed from the KV-cache, reducing memory consumption.
% The KV-cache retains only the KV vectors of the query prompt tokens and those of the previously generated tokens in the current reasoning step.
\textbf{In the SA submodule}, the KV‑cache retains only the KV vectors of the query prompt tokens and ones of the previously generated tokens in the current reasoning step, while those of tokens in completed reasoning steps have been removed.
Through the softmax attention mechanism (seen Eq.~\ref{q:1}), the input token $x_{t,i-1}$ attends to tokens retained in the KV‑cache, yielding an updated feature $\hat{h}_{t,i-1}^{(l)}$.
% Thus, the input token $x_{t,i-1}$ captures relevant information from these tokens using softmax attention (seen Eq.~\ref{q:1}), producing the updated feature $\hat{h}_{t,i-1}^{l-1}$.

\textbf{In the LA submodule}, after completing the first $t{-}1$ reasoning steps, our model state is updated to $\boldsymbol{S}_{t-1}^{(l)}$.
% $\boldsymbol{S}_{t-1}^l{=}\sum_{i'=1}^{T_{t-1}} {\boldsymbol{k}_{i'}^{l^{\top}}} \boldsymbol{v}_{i'}^l$ (seen Eq.~\ref{q:2}), where $T_{t-1}{=}| \boldsymbol{Q}|{+}\sum_{j=1}^{t-1}n_{j}$, $| \boldsymbol{Q}|$ and $n_j$ denote the number of tokens in the query prompt and in the $j$-th reasoning step, respectively.
Then, we use this state as the initial state $\boldsymbol{S}_{t,0}^{(l)}{=}\boldsymbol{S}_{t-1}^{(l)}$ for the current reasoning step and update it token‑by‑token, yielding the current model state $\boldsymbol{S}_{t,i-1}^{(l)}{=} \boldsymbol{S}_{t,0}^{(l)} {+} \sum_{j=1}^{i-1} {\boldsymbol{k}_{j}^{{(l)}^{\top}}} \boldsymbol{v}_{j}^{(l)}$.
Next, we utilize the query vector $\boldsymbol{q}_{i-1}^{(l)}$ of the input token $x_{i-1}$ to extract the relevant historical reasoning information $\boldsymbol{o}_{i-1}^{(l)}$ from $\boldsymbol{S}_{t,i-1}^{(l)}$: $\boldsymbol{o}_{i-1}^{(l)}{=}\boldsymbol{q}_{i-1}^{(l)} \boldsymbol{S}_{t,i-1}^{(l)}$. 
In the current reasoning step, the model usually relies more on historical reasoning information to generate token in the early stage, and this dependence gradually decreases as more tokens are generated.
Therefore, we obtain the output of this submodule via a gating mechanism: $\check{h}_{t,i-1}^{(l)}{=}\sigma(\boldsymbol{W}_g h_{t,i-1}^{(l-1)})*\boldsymbol{o}_{i-1}^{(l)}$, where $\sigma(\cdot)$ denotes the sigmoid function and $\boldsymbol{W}_g$ is a learnable weight (see Figure~\ref{fig2}).

Finally, we combine the outputs of the two submodules and apply a linear output layer to yield the final output of MAM: $\tilde{h}_{t,i-1}^{(l)}{=}\boldsymbol{W}_o(\check{h}_{t,i-1}^{(l)}{+}\hat{h}_{t,i-1}^{(l)})$.
As shown in Figure~\ref{fig2}, the two submodules have identical structures, except that the LA submodule incorporates an additional gating weight $\boldsymbol{W}_g$. 
To control its parameter size, we implement the LA submodule via the LoRA strategy \citep{c:45}.

Subsequently, we process $\tilde{h}_{t,i-1}^{(l)}$ using the FFN module to produce the output $h_{t,i-1}^{(l)}$ for the $l$-th layer of our model.
By repeating the above process $L$ times, we obtain the final output feature $h_{t,i-1}^{(L)}$ of the input token $x_{t-1}$, where $L$ denotes the number of layers in our model.
Next,  $h_{t,i-1}^{(L)}$ is fed into the model’s prediction layer to produce the predicted distribution $p(x)$ of the next token.
Finally, our model generates the $i$-th token $x_i$ of the $t$-th inference step according to $x_i {=} \arg\max_{x} p(x)$.

Following the above procedure, our model sequentially generates tokens in the current reasoning step until reaching its end token $\left \langle \setminus e_{t} \right \rangle$, which corresponds to the thinking pattern of the step (see Figure~\ref{fig1}).
After generating $\left \langle \setminus e_{t} \right \rangle$, the state matrix of our model at the $l$-th layer is updated to $\boldsymbol{S}_{t,n_t}^{(l)}$, which we denote as $\boldsymbol{S}_{t}^{(l)}$ and use as the initial state for the next reasoning step, where $n_i$ denotes the number of tokens in the $i$-th reasoning step.
Meanwhile, we clear the KV vectors of all tokens in the current inference step from the KV‑cache.

\subsection{The state-based reasoning strateg}
\label{S:3.2}

\begin{figure}[t]
\centering
\includegraphics[width=0.9 \textwidth]{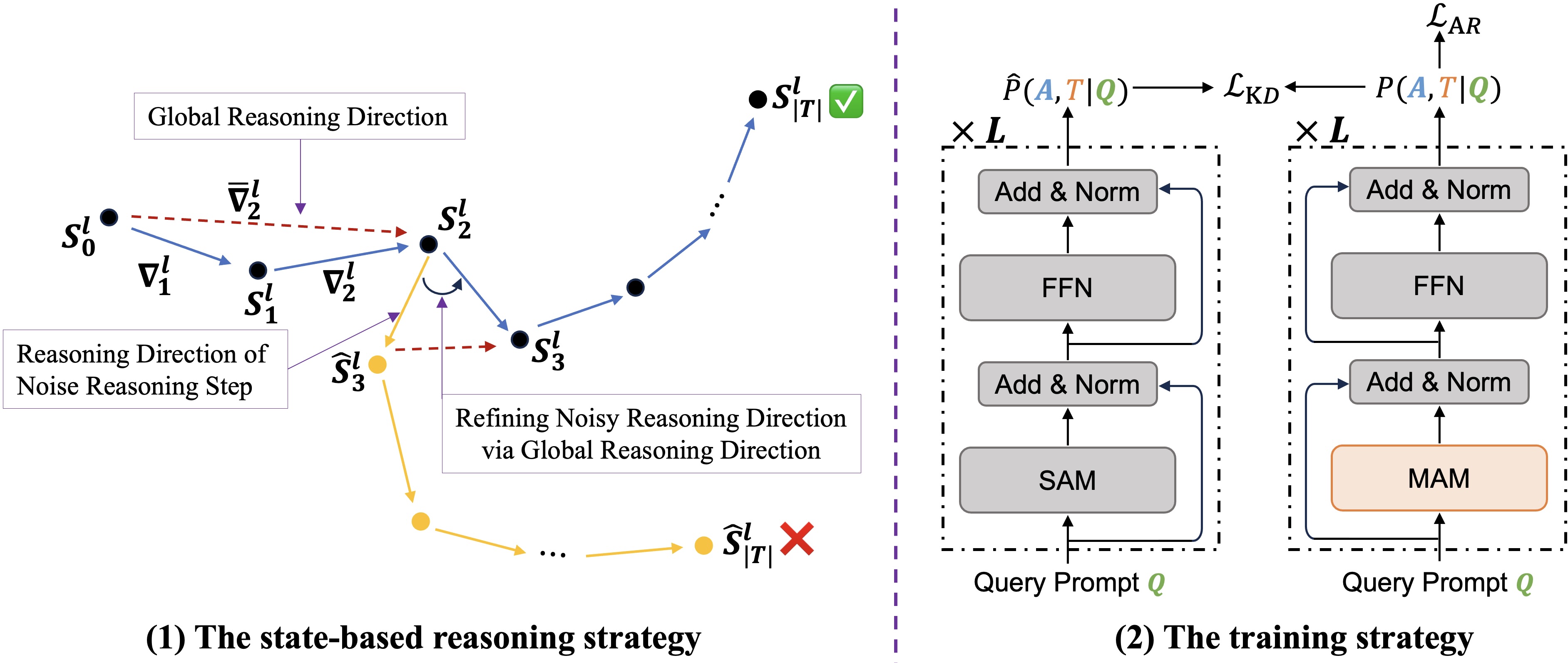}
\caption{Illustration of our reasoning and training strategies. SAM is the softmax attention module.
}
\label{fig3}
\end{figure}

During reasoning, LLMs often produce noisy reasoning steps that may mislead subsequent ones, thus resulting in overthinking problems \citep{c:33,c:34}.
In our framework, such noisy reasoning step can deviate the model’s state transitions from the correct reasoning trajectory, resulting in erroneous results (see Figure~\ref{fig3}(a)).
To mitigate this issue, we propose a state‑based reasoning strategy, which guides model reasoning with a global reasoning direction.
% to  reduce the bias from noisy reasoning steps.

In the reasoning process, the state transition of our model at the $l$-th layer can be formalized as: $\boldsymbol{S}_{0}^l {\to} \boldsymbol{S}_{1}^l {\to} \cdots {\to} \boldsymbol{S}_{|\boldsymbol{T}|}^l$, where $\boldsymbol{S}_{0}^l$ denotes the model state after the LA submodule processes the token in the query prompt $\boldsymbol{Q}$.
Considering that the state transition process in the line attention is a gradient descent process (see Section~\ref{s:2.1}), we represent the total gradient contributed by each reasoning step as $[\nabla_{1}^{l},\nabla_{2}^{l},\cdots,\nabla_{{|\boldsymbol{T}|}}^{l}]$, where $\nabla_{t}^{l}{=}\boldsymbol{S}_{t}^l{-}\boldsymbol{S}_{t-1}^l$ indicates the reasoning direction of the $t$-th step.
The reasoning direction of a noisy reasoning step often deviates substantially from those of other steps.

Therefore, we first aggregate the reasoning directions of all previous steps into a global reasoning direction $\bar{\nabla}_{t-1}^{l}$ using momentum accumulation: $\bar{\nabla}_{t-1}^{l}{=}(1{-}\frac{1}{t-1})\bar{\nabla}_{t-2}^{l}{+}\frac{1}{t-1}\nabla_{t-1}^{l}{=} \frac{1}{t-1}\sum_{i=1}^{t-1}\nabla_{i}^{l}$, where $\bar{\nabla}_{0}^{l}$ is initialized as a zero matrix.
Then, once the $t$-th reasoning step is completed, we employ the global direction $\bar{\nabla}_{t-1}^{l}$ to correct its reasoning direction $\nabla_{t}^{l}$: $\hat{\nabla}_{t}^{l}{=}(1{-}\alpha) \nabla_{t}^{l} {+}  \alpha\bar{\nabla}_{t-1}^{l}$, where $\alpha{=}\max\{\alpha_{\rm max},\frac{t}{|\boldsymbol{T}|}\}$ and $|\boldsymbol{T}|$ denotes the maximum number of reasoning steps (default: 40).
Since more prior reasoning steps yield a more accurate global reasoning direction, we linearly increase the correction coefficient $\alpha$ up to a predefined threshold $\alpha_{\rm max}$.
In this way, our model can fully explore diverse reasoning directions during the early stages of reasoning and then gradually converge toward the global reasoning direction.
Finally, we use the corrected reasoning direction $\hat{\nabla}_{t}^{l}$ to update the model state, $\boldsymbol{S}_{t}^l {=} \boldsymbol{S}_{t-1}^l{+}\hat{\nabla}_{t}^{l}$, alleviating the negative impact of the noisy step on subsequent steps.

To enhance the diversity of thinking patterns during reasoning, we apply special markers indicating thinking patterns to guide the model toward adopting different thinking patterns across consecutive steps.
By doing so, we can further enhance the robustness and overall performance of our model.

\subsection{Model training}
\label{S:3.3}

We train our model using the long CoT training set constructed in Section~\ref{s:2.1}.
To improve training efficiency while preserving the original reasoning ability of LLMs, we fine‑tune only the parameters of the newly added LA submodule and ones of the special tokens corresponding to thinking patterns.
As shown in Figure~\ref{fig3}(b), we jointly optimize our model with two loss terms: (1) the autoregressive loss $\mathcal{L}_{\rm AR}$ of our model on the training samples, and (2) the knowledge distillation loss $\mathcal{L}_{\rm KD}$ between the base model (the original LLM with softmax attention module) and our proposed model.
Finally, our training objective is defined as $\mathcal{L} {=} \mathcal{L}_{\rm AR}{+}\beta\mathcal{L}_{\rm KD}$, where $\beta$ denotes a hyperparameter.
% is a weight coefficient that balances the contribution of $\mathcal{L}_{\rm KD}$.

Specifically, we first input a long CoT sample into our model to obtain the predicted probability distribution $P(\boldsymbol{A},\boldsymbol{T}|\boldsymbol{Q})$ over the thinking sequence $\boldsymbol{T}$ and the final answer $\boldsymbol{A}$ conditioned on the query prompt $\boldsymbol{Q}$.
To maintain the consistency between training and testing, we apply a customized mask matrix in the SA submodule to ensure that each token attends only the tokens in the query prompt $\boldsymbol{A}$ and those from its corresponding reasoning step.
Next, our autoregressive loss term $\mathcal{L}_{\rm AR}$ is defined as follows:
$\mathcal{L}_{\rm AR} {=} -\log P(\boldsymbol{A},\boldsymbol{T}|\boldsymbol{Q})$.

Meanwhile, the same long CoT sample is fed into the base model to produce the predicted probability distribution $\hat{P}(\boldsymbol{A},\boldsymbol{T}|\boldsymbol{Q})$.
Notably, our model is built upon the base model by replacing its softmax attention module with our MAM.
In the base model, each token attends to all previous tokens. 
Subsequently, we use the Kullback–Leibler (KL) divergence between $P(\boldsymbol{A},\boldsymbol{T}|\boldsymbol{Q})$ and $\hat{P}(\boldsymbol{A},\boldsymbol{T}|\boldsymbol{Q})$  as our distillation loss term $\mathcal{L}_{\rm KD}{=} \textbf{KL}(\hat{P}(\boldsymbol{A},\boldsymbol{T}|\boldsymbol{Q})||P(\boldsymbol{A},\boldsymbol{T}|\boldsymbol{Q}))$.
The $\mathcal{L}_{\rm KD}$ loss term is designed to effectively train our LA submodule for capturing global reasoning information.

\section{experiments}
\subsection{Experiment Settings}
\label{sec:5.1}

\textbf{Implementation Details.}
Our experiments are primarily conducted on the Qwen-2.5 series models \citep{c:11}.
Meanwhile, we extract 95K high‑quality samples from OpenR1‑Math‑220K to construct our training set using the method described in Section~\ref{s:2.1}.
We initialize Qwen-2.5 using its corresponding distilled version of DeepSeek‑R1 \citep{c:7}, and then train it on the training set to obtain \textbf{our base model}.
Finally, our framework is built upon this base model.

\textbf{Baselines.}
We compare our framework with two types of baselines: efficient reasoning methods and KV‑cache reduction methods.
In the first category, \textbf{LightThinker} \citep{c:10} uses learnable special tokens to compress the reasoning information of completed  reasoning steps, improving the reasoning efficiency of LLMs.
\textbf{INFTYTHINK} \citep{c:50} compresses the information from previous reasoning steps into a concise summary.
In the second category, \textbf{H2O} \citep{c:51} reduces KV-cache by only retaining the KV vectors for a small set of heavy-hitter tokens and recent tokens with a greedy eviction strategy.
\textbf{SapLLM} \citep{c:52} only stores the KV vectors of the initial, neighboring, and separator tokens into the KV-cache of LLMs.
These baselines are trained with LoRA and use the same number of trainable parameters as our model.

\textbf{Dataset \& Metric.}
We evaluate our framework on seven benchmarks, including five mathematical reasoning benchmarks (GSM8K \citep{c:53}, MATH‑500 \citep{c:54}, AMC23 \citep{amc2023}, AIME24 \citep{aime2024}, AIME25 \citep{aime2025}), a scientific reasoning benchmark (GPQA\_Diamond \citep{c:55}), and a code reasoning benchmark (HumanEval \citep{c:56}).
We use greedy decoding to generate the output of the target LLM.
We report three metrics to assess the performance and efficiency of various methods: 
(1) \textit{Accuracy} (\textbf{Acc}) denotes the percentage of correct answers generated by the target model;
(2) \textit{Token Number} (\textbf{Tok}) refers to the average length of the CoT sequence;
(3) \textit{Reasoning Latency} (\textbf{ReL}) is defined as the average inference time per sample.
Moreover, we measure reasoning latency on a single NVIDIA A100 GPU with a batch size of $1$, while enabling FlashAttention-2 \citep{c:57} with BF16 \citep{c:58}.

\begin{table*}[t]
\setlength{\abovecaptionskip}{6pt}
\setlength{\belowcaptionskip}{-3pt}
\centering
\setlength{\tabcolsep}{1.8pt}
\resizebox{\textwidth}{!}{
\begin{tabular}{lccccccccccccccccc}
\toprule[1.2pt]
\multirow{2}{*}{\textbf{Method}}
 & \multicolumn{3}{c}{\textbf{GSM8K}} 
 & \multicolumn{3}{c}{\textbf{MATH-500}}
 & \multicolumn{3}{c}{\textbf{AMC23}}
 & \multicolumn{3}{c}{\textbf{AIME24}}
 & \multicolumn{3}{c}{\textbf{AIME25}}
 & \multicolumn{2}{c}{\textbf{AVG.}} \\
\cmidrule(lr){2-4} \cmidrule(lr){5-7} \cmidrule(lr){8-10} \cmidrule(lr){11-13} \cmidrule(lr){14-16} \cmidrule(lr){17-18}
 & Acc$\uparrow$ & Tok & ReL$\downarrow$ 
 & Acc$\uparrow$ & Tok & ReL$\downarrow$  
 & Acc$\uparrow$ & Tok & ReL$\downarrow$  
 & Acc$\uparrow$ & Tok & ReL$\downarrow$  
 & Acc$\uparrow$ & Tok & ReL$\downarrow$
 & Acc$\uparrow$ & ReL$\downarrow$ \\
\midrule[0.9pt]
\multicolumn{18}{c}{\textit{Qwen2.5-1.5B Series}} \\
\midrule[0.9pt]
\rowcolor{lightgray}Ours (Base)         &80.1 &2086 &76.2	&78.8 &3958 &139.4	&62.5 &6392 &242.9	&20.0 &13765 &536.8	&16.7 &12684 &504.7	&51.5 &300.0   \\
\;\;$+$H2O          &76.1 &2116 &71.7	&75.4 &3835 &128.5	&55.0 &6230 &209.7	&13.3 &13921 &462.2	&10.0  &12802  &428.6	    &46.0 &260.1  \\
\;\;$+$SepLLM       &77.3 &2230 &75.8	&76.0 &3890 &130.3	&57.5 &6573 &220.2	&13.3 &13690 &452.6	&13.3 &12690 &423.9	&47.5 &260.6 \\
LightThinker        &78.8 &2109 &69.6	&77.6 &4060 &134.8	&60.5 &6312 &214.6	&16.7 &13827 &461.6	&13.3 &12934  &434.6	&50.1 &263.0  \\
INFTYTHINK          &79.5 &2503 &89.3	&78.0 &4750 &170.2	&62.5 &7605 &263.5	&16.7 &16318 &566.2	&16.7 &15020 &501.0	&50.7 &318.0 \\
\rowcolor{lightblue}
% \rowcolor{blue!16}
\textbf{Ours}       &\textbf{82.1} &\textbf{2116} &\textbf{69.7}	&\textbf{81.2} &\textbf{3812} &\textbf{125.8}	&\textbf{67.5} &\textbf{6460} &\textbf{213.2}	&\textbf{26.7} &\textbf{13610} &\textbf{440.13}	&\textbf{23.3} &\textbf{12910} &\textbf{416.0}	&\textbf{56.2} &\textbf{253.0} \\
\midrule[0.9pt]
\multicolumn{18}{c}{\textit{Qwen2.5-7B Series}} \\
\midrule[0.9pt]
\rowcolor{lightgray}Ours (Base)          &89.4 &2649 &96.4	&87.4 &4253 &161.6	&82.5 &6704 &242.9	&40.0 &12901 &522.1	&30.0 &13204 &548.1	&65.9 &314.2   \\
\;\;$+$H2O           &83.5 &2730 &92.0	&82.6 &4389 &147.1	&77.5 &6843 &209.7	&30.0 &13274 &448.9	&23.3 &13204 &443.7	&59.4 &268.3  \\
\;\;$+$SepLLM        &84.9 &2582 &92.3	&84.2 &4244 &144.3	&80.0 &6511 &220.2	&33.3 &13100 &484.7	&23.3 &12972 &435.9	&61.1 &275.5 \\
LightThinker         &85.6 &2779 &90.0	&84.8 &4321 &145.2	&80.0 &6790 &214.6	&33.3 &13307 &449.0	&26.7 &13250 &443.9	&62.1 &268.5  \\
INFTYTHINK           &87.9 &3237 &117.0	&86.2 &5050 &184.8	&82.5 &8061 &254.3	&36.7 &15094 &508.7	&30.0 &15448 &530.5	&64.7 &319.0 \\
\rowcolor{lightblue}
% \rowcolor{blue!16}
\textbf{Ours}                 &\textbf{90.9} &\textbf{2603} &\textbf{87.2}	&\textbf{90.0} &\textbf{4196} &\textbf{140.6}	&\textbf{85.0} &\textbf{6665} &\textbf{213.2}	&\textbf{43.3} &\textbf{13017} &\textbf{434.8}	&\textbf{33.3} &\textbf{13191} &\textbf{440.6}	&\textbf{68.3} &\textbf{263.3} \\
\bottomrule[1.2pt]
\end{tabular}
}
\caption{The results of our model and baselines on mathematical reasoning benchmarks.}
\label{table:1}
\end{table*}

\subsection{Main Results}
The experimental results on mathematical benchmarks are presented in Table~\ref{table:1}.
As shown in the \textit{AVG.} column, our framework outperforms all baselines in reasoning efficiency and attains the best overall performance.
% our framework not only outperforms all baselines in reasoning efficiency but also attains the best overall performance.
After carefully analyzing these results, we draw several conclusions:

\textbf{First}, our model significantly outperforms the base model in reasoning speed, particularly on benchmarks requiring longer CoT.
On the AIME24 benchmark, our model achieved a $16{-}18\%$ improvement in reasoning speed over the base model.
This is mainly attributed to the lower computational complexity of our MAM relative to the softmax attention module of the base model.
Furthermore, our model consistently achieves superior performance over the base model across all benchmarks.
These results highlight the strong potential of linear attention mechanisms for efficient reasoning.

\textbf{Second}, the baselines (except INFTYTHINK) and our model generate CoT sequences with similar length on each benchmark, as they are trained on the same dataset.
While these baselines achieve higher reasoning efficiency than the base model, they exhibit worse reasoning performance.
For instance, on the AIME25 benchmark, LightThinker improves reasoning efficiency by $10{–}12\%$ over the base model, but its performance decreases by $3{–}6$ points.
These results suggest that naive KV‑cache compression methods risk losing important reasoning information, thus limiting model performance.
% In contrast, our model does not suffer from this issue.

\textbf{Third}, unlike other baselines, INFTYTHINK uses text summaries of previous reasoning steps to record their reasoning information.
While enhancing interpretability, this method reduces reasoning efficiency because it requires generating longer CoT sequences.
Moreover, such discrete summarization may omit part of the implicit reasoning information within the model, leading to a slight performance drop for INFTYTHINK compared to the base model.
However, we effectively avoid these two issues by utilizing the model’s reasoning states to record reasoning information, enabling our model to outperform INFTYTHINK in both efficiency and performance.

\begin{table}[t]
\setlength{\abovecaptionskip}{6pt}
\setlength{\belowcaptionskip}{-3pt}
\centering
\small
\setlength{\tabcolsep}{2.5pt}
\renewcommand{\arraystretch}{0.7}
\begin{tabular}{lcccccc}
\toprule[1.2pt]
\textbf{Method} & \textbf{GSM8K} & \textbf{MATH-500} & \textbf{AMC23} & \textbf{AIME24} & \textbf{AIME2}5 & \textbf{AVG.} \\ \rowcolor{lightblue}
\midrule[0.8pt]
Ours   &\textbf{82.1} 	&\textbf{81.2}	 &\textbf{67.0} 	&\textbf{26.7} 	&\textbf{23.3}	&\textbf{56.2} \\
\midrule[0.6pt]
\;\;\;\;(1) \textit{w/o} The state‑based reasoning strategy         &79.3	&78.8	&65.0	&20.0	&20.0	&52.6 \\
\;\;\;\;(2) \textit{w/o} Thinking patterns diversity    &80.9	&80.0	&65.0	&23.3	&20.0	&53.8 \\
\;\;\;\;(3) \textit{w/o} The distillation loss $\mathcal{L}_{\rm KD}$      &77.0	&77.6	&62.5	&20.0	&16.7	&50.8 \\
\;\;\;\;(4) \textit{w/o} The model reasoning state  &76.4	&76.2	&62.5	&20.0	&16.7	&46.8 \\
\;\;\;\;(5) \textit{w/o} The gating mechanism          &80.1	&79.4	&65.0	&23.3	&23.3	&54.2 \\
\bottomrule[1.2pt]
\end{tabular}
\caption{Ablation results of our model on mathematical reasoning benchmarks.}
\label{table:2}
\end{table}

\subsection{Ablation Study}
We further conduct extensive ablation studies by removing different components from our framework to investigate their different impacts. 
As shown in Table~\ref{table:2}, we only compare our model with the following variants in terms of performance, as they share the same reasoning efficiency.

\textit{w/o The state‑based reasoning strategy.} 
In this variant, we remove the state‑based reasoning strategy from our framework.
We observe an average performance drop of 3.54 points across all benchmarks for this variant.
These results suggest that our reasoning strategy can indeed enhance the robustness of LLMs to noisy reasoning steps, improving their overall performance.
Moreover, this also indicates that, with suitable processing, our model states can intuitively reflect the quality of reasoning steps, which offers an interesting insight for future exploration (e.g., process-reward-based RL).

\textit{w/o Thinking patterns diversity.} 
We guide LLMs to use different thinking patterns in adjacent reasoning steps via corresponding special tokens, thus ensuring diversity of the thinking patterns in the reasoning process.
To test its effectiveness, we remove this operation from our framework.
As shown in \textbf{Line (2)}, this variant exhibits a consistent decline in performance across all benchmarks.
These findings imply that more diverse thinking patterns can enhance LLMs’ exploration ability, thereby improving the accuracy of their generated answers.

\textit{The distillation loss $\mathcal{L}_{\rm KD}$.} 
As described in Section~\ref{S:3.3}, we use $\mathcal{L}_{\rm KD}$ to train our LA submodule so that it can better capture global reasoning information.
In this variant, $\mathcal{L}_{\rm KD}$ is excluded from our training objective.
From \textbf{Line~(3)} in Table~\ref{table:2}, we note that this variant results in a performance drop of $3.6{–}6.6$ points across all benchmarks.
The main reason is that the autoregressive loss $\mathcal{L}_{\rm AR}$  may struggle to train the LA submodule to capture the global correlations between reasoning steps.

\textit{\textit{w/o} The model reasoning state.} 
In this variant, we replace the model states in our framework with a sliding window of size 64 to store historical reasoning information.
Although this approach is also compatible with the original softmax attention in LLMs, this variant yields the most significant performance drop compared with other variants.
This is mainly because this approach lacks reasoning capability as strong as that of linear attention.

\textit{\textit{w/o} The gating mechanism.} 
This variant removes the gating mechanism from our LA submodule and directly sums the outputs of the two submodules in the MAM.
This change leads to a slight performance drop across all benchmarks.
This confirms that our gating mechanism can more precisely control the amount of historical reasoning information each token requires.

\subsection{Further Analysis}
\begin{figure}[t]
\centering
\includegraphics[width=1.0 \textwidth]{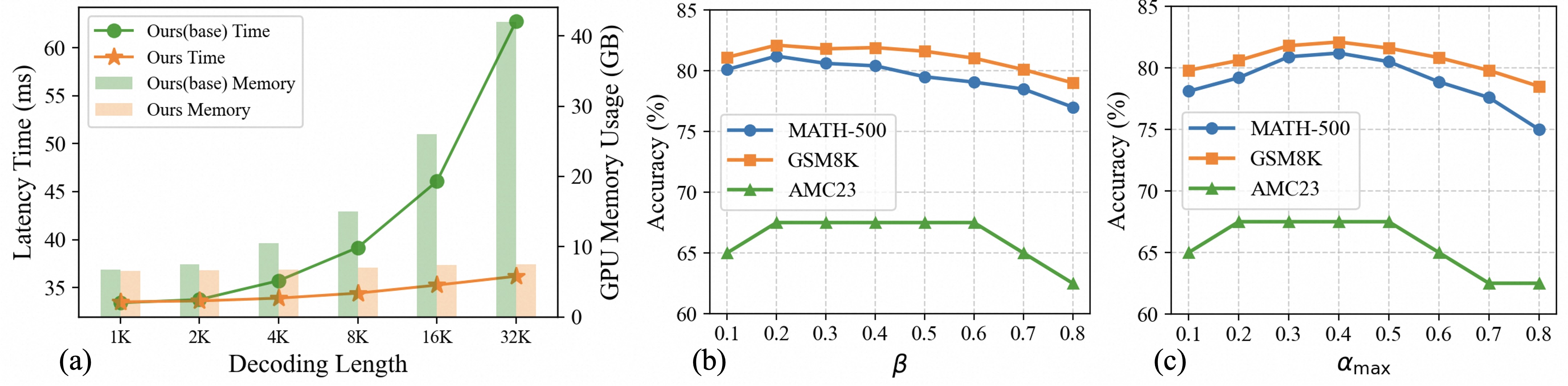}
\caption{(a) shows the computational and memory efficiency of our model and the base model.
(b) and (c) present our model’s performance with different values of hyper-parameters $\beta$ and $\alpha_{\rm max}$, respectively.
These experiments are conducted on Qwen2.5-1.5B.
}
\label{fig4}
\end{figure}

\textbf{Analysis of Computational and Memory Costs.}
We conduct experiments to further compare the computational and memory efficiency of our model and the base model across varying CoT lengths.
The experimental results are presented in Figure~\ref{fig4}(a).
Although our model exhibits similar reasoning efficiency to the base model for shorter CoT, it significantly surpasses the base model once the CoT length exceeds 4K.
In particular, when the CoT length reaches 32K, our model achieves over 40\% faster reasoning speed than the base model.
Moreover, our model maintains a nearly constant memory usage across varying CoT lengths, whereas that of the base model increases linearly with CoT length.
Theoretically, our model’s advantages in computational and memory efficiency would become even more significant when FlashAttention-2 is disabled.

\textbf{Analysis of Hyper-Parameters.}
We also investigate the impact of the two key hyper-parameters, $\beta$ and $\alpha_{\rm max}$, on the performance of our model.
As illustrated in Figure~\ref{fig4}(b)–(c), our model exhibits low sensitivity to these two hyper-parameters.
Meanwhile, our model attains the best performance when $\beta$ and $\alpha_{\rm max}$ are set to 0.2 and 0.4, respectively. 
We further analyze the choice of these two hyperparameter values as follows:
\begin{itemize}
  \item $\beta$ denotes the weight of the distillation loss term $\mathcal{L}_{\rm KD}$. Setting $\beta$ to an excessively small value (e.g., 0.1) prevents our linear attention module from effectively capturing global reasoning information, whereas an overly large value (e.g., 0.8) limits the potential of our model to outperform the base model.
  \item In the state‑based reasoning strategy, $\alpha$ determines the magnitude of correction applied to the reasoning direction of noisy reasoning steps. Hence, setting $\alpha$ too low (e.g., 0.1) increases the model’s sensitivity to noisy reasoning steps, whereas setting it too high (e.g., 0.8) constrains the exploration capabilities of LLMs.
  \item Moreover, within an appropriate range (e.g., 0.2$-$0.6), our model’s performance is insensitive to these hyperparameters.
\end{itemize}

\textbf{Analysis of More Baselines.}
Recently, many studies \citep{c:59,c:60,c:61,c:62,c:63,c:64,c:77,c:65,c:66,c:67,c:68,c:69,c:70,c:78,c:71,c:72,c:73,c:74,c:76} have focused on shortening the length of CoT to improve the reasoning efficiency of LLMs.
However, our method aims to reduce both the computational and storage complexity of attention in LLMs, leading to faster token generation.
This indicates that our method improves the reasoning efficiency of LLMs regardless of the CoT length.
Therefore, these baselines are both orthogonal to and compatible with our method.
Here, we select four open-source baselines as examples to verify the compatibility of our method with these baselines: Task Arithmetic-based Model Merging(\textbf{TAMM}) \citep{c:60}, \textbf{SBT-D} \citep{c:72}, \textbf{L1-Max} \citep{c:61}, and  Anytime-Reasoning (\textbf{AR}) \citep{c:74}.
As shown in Table~\ref{table:1-1}, our method further enhances the reasoning efficiency of these baselines by accelerating the token generation speed of LLMs.
Furthermore, RL‑based baselines (e.g., AR) provide stronger base models for our method, thus further raising its performance upper bound.

\begin{table*}[t]
\setlength{\abovecaptionskip}{6pt}
\centering
\setlength{\tabcolsep}{1.8pt}
\resizebox{\textwidth}{!}{
\begin{tabular}{lccccccccccccccccc}
\toprule[1.2pt]
\multirow{2}{*}{\textbf{Method}}
 & \multicolumn{3}{c}{\textbf{GSM8K}} 
 & \multicolumn{3}{c}{\textbf{MATH-500}}
 & \multicolumn{3}{c}{\textbf{AMC23}}
 & \multicolumn{3}{c}{\textbf{AIME24}}
 & \multicolumn{3}{c}{\textbf{AIME25}}
 & \multicolumn{2}{c}{\textbf{AVG.}} \\
\cmidrule(lr){2-4} \cmidrule(lr){5-7} \cmidrule(lr){8-10} \cmidrule(lr){11-13} \cmidrule(lr){14-16} \cmidrule(lr){17-18}
 & Acc$\uparrow$ & Tok & ReL$\downarrow$ 
 & Acc$\uparrow$ & Tok & ReL$\downarrow$  
 & Acc$\uparrow$ & Tok & ReL$\downarrow$  
 & Acc$\uparrow$ & Tok & ReL$\downarrow$  
 & Acc$\uparrow$ & Tok & ReL$\downarrow$
 & Acc$\uparrow$ & ReL$\downarrow$ \\
\midrule[0.9pt]
\multicolumn{18}{c}{\textit{Qwen2.5-1.5B Series}} \\
\midrule[0.9pt]
\rowcolor{lightgray}Ours (Base)         &80.1 &2086 &76.2	&78.8 &3958 &139.4	&62.5 &6392 &242.9	&20.0 &13765 &536.8	&16.7 &12684 &504.7	&51.5 &300.0   \\
\;\;$+$TAMM          &78.0 &1212 &43.4  &73.2 &2265 &82.1  &55.0 &2924 &105.6    &16.7 &5702 &215.8  &16.7 &5010 &189.0  &47.9 &127.2  \\
\;\;$+$SBT-D   &77.8 &1170 &43.1 	&72.8 &2105 &76.7 	&57.5 &3215 &114.9 	&16.7 &6409 &246.4 	&16.7 &6173 &235.5 	&48.3  &143.3  \\
\rowcolor{lightblue}
Ours       &82.1 &2116 &69.7	&81.2 &3812 &125.8	&67.5 &6460 &213.2	&26.7 &13610 &440.13	&23.3 &12910 &416.0	&56.2 &253.0 \\
\;\;$+$TAMM        &82.1 &2116 &69.7	&81.2 &3812 &125.8	&67.5 &6460 &213.2	&26.7 &13610 &440.13	&23.3 &12910 &416.0	&56.2 &253.0 \\
\;\;$+$SBT-D       &79.9 &1081 &36.2 	&76.8 &2110 &70.3 	&62.5 &3193 &101.4 	&23.3 &6458 &212.6 	&20.0 &6106 &201.5 	&52.1 &124.4  \\
\midrule[0.9pt]
\rowcolor{lightgray} \textit{Token Budget}          & &2k &	& &4k &	& &6k & 	& &13k &	& &12k &	& &   \\
Base$+$L1-Max      &80.5 &2077 &77.4  &79.0 &4091 &144.8   &62.5 &6103 &233.5 &20.0 &13108 &514.5  &16.7 &11904 &475.5  &51.7 &289.1  \\
Base$+$AR      &79.8 &2061 &75.8 	&79.4 &4083 &143.6 	&62.5 &6097 &232.0	&23.3 &13108 &514.5 	&20.0 &12088 &483.8 	&53.0  &289.9 \\
Ours$+$L1-Max    &82.7 &2103 &69.5  &81.4 &3962 &129.1  &65.0 &6025 &199.2  &26.7 &12980 &415.2  &23.3 &11871 &382.0  &55.8 &239.0  \\
Ours$+$AR       &82.7  &2084  &68.6 	&82.0  &4079  &131.2 	&67.5  &6101  &200.5 	&30.0 &12980  &415.2 	&26.7  &12101  &389.3 	&57.8  &240.9  \\
\midrule[0.9pt]
\rowcolor{lightgray} \textit{Token Budget}          & &1k &	& &2k &	& &3k & 	& &6.5k &	& &6k &	& &   \\
Base$+$L1-Max     &77.8 &1161 &41.6  &76.8 &2171 &80.3  &57.5 &3206 &124.6 &16.7 &6606 &260.7  &16.7 &6075 &230.8  &49.1 &147.6 \\
Base$+$AR      &78.8 &1105  &39.8 	&78.2  &2075  &77.8 	&60.0  &3112  &124.7	&20.0  &6587  &260.3 	&16.7  &6094  &231.6 	&50.7  &146.8 \\
Ours$+$L1-Max     &79.4 &1130 &36.3  &79.2  &2066  &68.6   &62.5  &3147  &103.1  &23.3 &6691 &208.0  &23.3  &6030 &198.1  &53.5  &122.8  \\
Ours$+$AR      &81.9 &1092  &35.1 	&80.4   &2107  &69.1 	 &65.0  &3107 &101.6 	&26.7  &6603  &207.4 	&23.3  &6088  &198.8 	 &55.5 &122.4  \\
\bottomrule[1.2pt]
\end{tabular}
}
\caption{The results of our model combined with baselines on mathematical reasoning benchmarks.}
\label{table:1-1}
\end{table*}

\textbf{In Appendix~\ref{appendix:A}}, we further analyze both the domain generalization capability of our framework and the gating mechanism within our LA submodule. 
\textbf{In Appendix~\ref{appendix:F}}, we present a more intuitive explanation of our framework and further examine its potential.

\section{Related work}
\textbf{Efficient Reasoning.}
Although long CoT enhances the reasoning ability of LLMs, it introduces significant computational overhead.
To alleviate this challenge, existing methods for CoT compression can be roughly divided into three categories. 
\textbf{The first category} leverages prompting \citep{c:12,c:14}, supervised fine‑tuning (SFT) \citep{c:15,c:25}, and reinforcement learning (RL) \citep{c:26,c:27,c:75} to encourage LLMs to generate shorter CoT sequences.
Since such CoT reductions conflict with test-time scaling \citep{c:6}, these methods often impair the reasoning ability of LLMs \citep{c:16,c:17}.
\textbf{The second category} compresses information from multiple text tokens into a single hidden vector, converting longer discrete CoT sequences into shorter continuous ones to improve the reasoning efficiency of LLMs \citep{c:20,c:19}.
However, such methods typically require altering the input/output patterns of LLMs, increasing their vulnerability to catastrophic forgetting and limiting their performance.
\textbf{The third category} expresses the CoT in more concise text to reduce its length while preserving the reasoning ability of LLMs \citep{c:23,c:22,c:21}.
Nevertheless, when simplifying long CoT, these methods risk losing critical reasoning information or reducing interpretability \citep{c:24}.
% In contrast, our framework enhances the reasoning efficiency of LLMs by reducing the computational complexity of their attention, without altering the CoT sequences.
% This preserves both the reasoning capability of LLMs and the interpretability of CoT sequences.

% To alleviate this challenge, existing methods for CoT compression can be broadly divided into three categories.
% The first category encourages LLMs to generate shorter CoT sequences via prompting \citep{c:12,c:14}, SFT \citep{c:15,c:25}, or RL \citep{c:26,c:27}.
% The second category focuses on compressing long discrete CoT sequences into shorter continuous ones, enhancing the reasoning efficiency of LLMs \citep{c:20,c:19}.
% The third category aims to shorten CoT by expressing it in more concise text, thus improving the reasoning efficiency of LLMs \citep{c:23,c:22,c:21}.
% While these methods yield improvements, shortening the CoT sequence conflicts with test-time scaling, limiting the reasoning capabilities of LLMs.
% Meanwhile, simplifying long CoT sequences may reduce their interpretability.
% In contrast, our framework enhances the inference efficiency of LLMs by reducing the computational complexity of attention, without altering the CoT sequences.
% This preserves both the reasoning capability of LLMs and the interpretability of CoT sequences.

\textbf{Linear Attention.}
To address the quadratic computational cost of softmax attention \citep{c:80,c:81}, many linear attention methods have been proposed to improve training and inference efficiency.
Early studies investigated kernel functions for linear attention \citep{c:38,c:39}.
Subsequently, some studies introduced gating mechanisms into linear attention to more effectively control the information in the model state \citep{c:46,c:47,c:48}.
Recent studies has provided theoretical evidence for the TTT property of linear attention \citep{c:31,c:41,c:49,c:33}.
They focus on exploring online learning objectives for linear attention to enhance its sequence modeling capability.
% Although the TTT property endows linear attention strong reasoning potential, its application in reasoning remains under-explored.

\section{Conclusion}
In this paper, we propose an efficient reasoning framework that models the reasoning process of LLMs as a state‑transition process to enhance their reasoning efficiency.
In our framework, we design an MAM to replace the original attention module in LLMs.
The MAM uses a linear attention mechanism to capture the model’s reasoning state that stores reasoning information from previous reasoning steps.
By utilizing the model state, we can significantly reduce the computational complexity and memory cost of attention, thereby improving the reasoning efficiency of LLMs.
Furthermore, we design a state-based reasoning strategy to avoid the overthinking issue caused by noisy reasoning steps.
Extensive experiments across multiple benchmarks and model sizes demonstrate the efficiency and robustness of our framework.

\section*{acknowledgements}
The project was supported by National Key R\&D Program of China (No. 2022ZD0160501), Natural Science Foundation of Fujian Province of China (No. 2024J011001), and the Public Technology Service Platform Project of Xiamen (No.3502Z20231043). We also thank the reviewers for their insightful comments.

\section*{Ethics Statement}
\label{appendix:C}
This work adheres to the ICLR Code of Ethics. In this study, no human subjects or animal experimentation was involved. All datasets used, including OpenR1-Math-220K, GSM8K, MATH-500, AMC23, AIME24, AIME25, GPQA\_Diamond, and HumanEval, were sourced in compliance with relevant usage guidelines, ensuring no violation of privacy. We have taken care to avoid any biases or discriminatory outcomes in our research process. No personally identifiable information was used, and no experiments were conducted that could raise privacy or security concerns. We are committed to maintaining transparency and integrity throughout the research process.

\section*{Reproducibility Statement}
\label{appendix:D}
We have made every effort to ensure that the results presented in this paper are reproducible. 
The experimental setup, including training steps, model configurations, and hardware details, is described in detail in the paper. 
Additionally, we use publicly available datasets, such as OpenR1-Math-220K, GSM8K, MATH-500, and HumanEval, to ensure consistent and reproducible evaluation results.
We believe these measures will enable other researchers to reproduce our work and further advance the field.

\section*{LLM Usage}
\label{appendix:E}
In the preparation of this paper, we use Large Language Models (LLMs) solely to aid in writing and polishing the text, including improving clarity, grammar, and readability. LLMs are not used for generating scientific content, experimental design, analysis, or conclusions. All technical ideas, experiments, and results reported in this paper are entirely the work of the authors.

\bibliography{iclr2026_conference}

@inproceedings{c:1,
  title={\href{https://proceedings.neurips.cc/paper/2022/hash/9d5609613524ecf4f15af0f7b31abca4-Abstract-Conference.html}{Chain-of-thought prompting elicits reasoning in large language models}},
  author={Wei, Jason and Wang, Xuezhi and Schuurmans, Dale and Bosma, Maarten and Xia, Fei and Chi, Ed and Le, Quoc V and Zhou, Denny and others},
  booktitle={Proceedings of NIPS},
  year={2022}
}

@inproceedings{c:2,
  title={\href{https://proceedings.neurips.cc/paper_files/paper/2022/hash/8bb0d291acd4acf06ef112099c16f326-Abstract-Conference.html}{Large language models are zero-shot reasoners}},
  author={Kojima, Takeshi and Gu, Shixiang Shane and Reid, Machel and Matsuo, Yutaka and Iwasawa, Yusuke},
  booktitle={Proceedings of NIPS},
  year={2022}
}

@inproceedings{c:3,
  title={\href{https://proceedings.neurips.cc//paper_files/paper/2023/hash/271db9922b8d1f4dd7aaef84ed5ac703-Abstract-Conference.html}{Tree of thoughts: Deliberate problem solving with large language models}},
  author={Yao, Shunyu and Yu, Dian and Zhao, Jeffrey and Shafran, Izhak and Griffiths, Tom and Cao, Yuan and Narasimhan, Karthik},
  booktitle={Proceedings of NIPS},
  year={2023}
}

@inproceedings{c:4,
  title={\href{https://openreview.net/forum?id=1PL1NIMMrw}{Self-consistency improves chain of thought reasoning in language models}},
  author={Wang, Xuezhi and Wei, Jason and Schuurmans, Dale and Le, Quoc and Chi, Ed and Narang, Sharan and Chowdhery, Aakanksha and Zhou, Denny},
  booktitle={Proceedings of ICLR},
  year={2023}
}

@inproceedings{c:5,
  title={\href{https://openreview.net/forum?id=WZH7099tgfM}{Least-to-most prompting enables complex reasoning in large language models}},
  author={Zhou, Denny and Sch{\"a}rli, Nathanael and Hou, Le and Wei, Jason and Scales, Nathan and Wang, Xuezhi and Schuurmans, Dale and Cui, Claire and Bousquet, Olivier and Le, Quoc and others},
  booktitle={Proceedings of ICLR},
  year={2023}
}

@article{c:6,
  title={\href{https://arxiv.org/abs/2412.16720}{Openai o1 system card}},
  author={OpenAI, Jaech, Aaron and Kalai, Adam and Lerer, Adam and Richardson, Adam and El-Kishky, Ahmed and Low, Aiden and Helyar, Alec and Madry, Aleksander and Beutel, Alex and Carney, Alex and others},
  journal={arXiv:2412.16720},
  year={2024}
}

@article{c:7,
  title={\href{https://arxiv.org/abs/2501.12948}{Deepseek-r1: Incentivizing reasoning capability in llms via reinforcement learning}},
  author={DeepSeek and Guo, Daya and Yang, Dejian and Zhang, Haowei and Song, Junxiao and Zhang, Ruoyu and Xu, Runxin and Zhu, Qihao and Ma, Shirong and Wang, Peiyi and Bi, Xiao and others},
  journal={arXiv:2501.12948},
  year={2025}
}

@article{c:8,
  title={\href{https://qwenlm.github.io/blog/qwq-32b-preview/}{Qwq: Reflect deeply on the boundaries of the unknown}},
  author={Team, Qwen},
  year={2024}
}

@inproceedings{c:9,
  title={\href{https://proceedings.neurips.cc/paper/2017/hash/3f5ee243547dee91fbd053c1c4a845aa-Abstract.html}{Attention is all you need}},
  author={Vaswani, Ashish and Shazeer, Noam and Parmar, Niki and Uszkoreit, Jakob and Jones, Llion and Gomez, Aidan N and Kaiser, {\L}ukasz and Polosukhin, Illia},
  booktitle={Proceedings of NIPS},
  year={2017}
}

@article{c:10,
  title={\href{https://arxiv.org/abs/2502.15589}{Lightthinker: Thinking step-by-step compression}},
  author={Zhang, Jintian and Zhu, Yuqi and Sun, Mengshu and Luo, Yujie and Qiao, Shuofei and Du, Lun and Zheng, Da and Chen, Huajun and Zhang, Ningyu},
  journal={arXiv:2502.15589},
  year={2025}
}

@article{c:11,
  title={\href{https://arxiv.org/abs/2412.15115}{Qwen2.5 Technical Report}},
  author={Hui, Binyuan and Yang, Jian and Cui, Zeyu and Yang, Jiaxi and Liu, Dayiheng and Zhang, Lei and Liu, Tianyu and Zhang, Jiajun and Yu, Bowen and Lu, Keming and others},
  journal={arXiv:2409.12186},
  year={2024}
}

@article{c:12,
  title={\href{https://arxiv.org/abs/2502.15589}{Token-budget-aware llm reasoning}},
  author={Han, Tingxu and Wang, Zhenting and Fang, Chunrong and Zhao, Shiyu and Ma, Shiqing and Chen, Zhenyu},
  journal={arXiv:2412.18547},
  year={2024}
}

@article{c:14,
  title={\href{https://arxiv.org/abs/2504.09858}{Reasoning models can be effective without thinking}},
  author={Ma, Wenjie and He, Jingxuan and Snell, Charlie and Griggs, Tyler and Min, Sewon and Zaharia, Matei},
  journal={arXiv:2504.09858},
  year={2025}
}

@inproceedings{c:15,
  title={\href{https://proceedings.neurips.cc/paper_files/paper/2024/hash/504fa7e518da9d1b53a233ed20a38b46-Abstract-Conference.html}{Can language models learn to skip steps?}},
  author={Liu, Tengxiao and Guo, Qipeng and Hu, Xiangkun and Jiayang, Cheng and Zhang, Yue and Qiu, Xipeng and Zhang, Zheng},
  booktitle={Proceedings of NIPS},
  year={2024}
}

@inproceedings{c:16,
  title={\href{https://aclanthology.org/2024.findings-acl.108/}{The impact of reasoning step length on large language models}},
  author={Jin, Mingyu and Yu, Qinkai and Shu, Dong and Zhao, Haiyan and Hua, Wenyue and Meng, Yanda and Zhang, Yongfeng and Du, Mengnan},
  booktitle={Proceedings of ACL},
  year={2024}
}

@inproceedings{c:17,
  title={\href{https://aclanthology.org/2024.findings-acl.108/}{The expressive power of transformers with chain of thought}},
  author={Merrill, William and Sabharwal, Ashish and Merrill, William and Sabharwal, Ashish},
  booktitle={Proceedings of ICLR},
  year={2024}
}

@article{c:19,
  title={\href{https://arxiv.org/abs/2502.03275}{Token assorted: Mixing latent and text tokens for improved language model reasoning}},
  author={Su, DiJia and Zhu, Hanlin and Xu, Yingchen and Jiao, Jiantao and Tian, Yuandong and Zheng, Qinqing},
  journal={arXiv:2502.03275},
  year={2025}
}

@article{c:20,
  title={\href{https://arxiv.org/abs/2412.06769}{Training large language models to reason in a continuous latent space}},
  author={Hao, Shibo and Sukhbaatar, Sainbayar and Su, DiJia and Li, Xian and Hu, Zhiting and Weston, Jason and Tian, Yuandong},
  journal={arXiv:2412.06769},
  year={2024}
}

@article{c:21,
  title={\href{https://arxiv.org/abs/2412.06769}{Tokenskip: Controllable chain-of-thought compression in llms}},
  author={Xia, Heming and Leong, Chak Tou and Wang, Wenjie and Li, Yongqi and Li, Wenjie},
  journal={arXiv:2502.12067},
  year={2025}
}

@inproceedings{c:22,
  title={\href{https://ojs.aaai.org/index.php/AAAI/article/view/34608}{C3ot: Generating shorter chain-of-thought without compromising effectiveness}},
  author={Kang, Yu and Sun, Xianghui and Chen, Liangyu and Zou, Wei},
  booktitle={Proceedings of AAAI},
  year={2025}
}

@article{c:23,
  title={\href{https://arxiv.org/abs/2502.09601}{Cot-valve: Length-compressible chain-of-thought tuning}},
  author={Ma, Xinyin and Wan, Guangnian and Yu, Runpeng and Fang, Gongfan and Wang, Xinchao},
  journal={arXiv:2502.09601},
  year={2025}
}

@article{c:24,
  title={\href{https://arxiv.org/abs/2505.16838}{R1-Compress: Long Chain-of-Thought Compression via Chunk Compression and Search}},
  author={Wang, Yibo and Shen, Li and Yao, Huanjin and Huang, Tiansheng and Liu, Rui and Tan, Naiqiang and Huang, Jiaxing and Zhang, Kai and Tao, Dacheng},
  journal={arXiv:2505.16838},
  year={2025}
}

@article{c:25,
  title={\href{https://arxiv.org/abs/2502.20122}{Self-training elicits concise reasoning in large language models}},
  author={Munkhbat, Tergel and Ho, Namgyu and Kim, Seo Hyun and Yang, Yongjin and Kim, Yujin and Yun, Se-Young},
  journal={arXiv:2502.20122},
  year={2025}
}

@article{c:26,
  title={\href{https://arxiv.org/abs/2503.04697}{L1: Controlling how long a reasoning model thinks with reinforcement learning}},
  author={Aggarwal, Pranjal and Welleck, Sean and Aggarwal, Pranjal and Welleck, Sean},
  journal={arXiv:2503.04697},
  year={2025}
}

@article{c:27,
  title={\href{https://arxiv.org/abs/2503.04472}{Dast: Difficulty-adaptive slow-thinking for large reasoning models}},
  author={Shen, Yi and Zhang, Jian and Huang, Jieyun and Shi, Shuming and Zhang, Wenjing and Yan, Jiangze and Wang, Ning and Wang, Kai and Liu, Zhaoxiang and Lian, Shiguo},
  journal={arXiv:2503.04472},
  year={2025}
}

@inproceedings{c:28,
  title={\href{https://proceedings.mlr.press/v119/katharopoulos20a.html?ref=mackenziemorehead.com}{Transformers are rnns: Fast autoregressive transformers with linear attention}},
  author={Katharopoulos, Angelos and Vyas, Apoorv and Pappas, Nikolaos and Fleuret, Fran{\c{c}}ois},
  booktitle={Proceedings of ICML},
  year={2020}
}

@inproceedings{c:29,
  title={\href{https://proceedings.neurips.cc/paper/2012/hash/c399862d3b9d6b76c8436e924a68c45b-Abstract.html}{Imagenet classification with deep convolutional neural networks}},
  author={Krizhevsky, Alex and Sutskever, Ilya and Hinton, Geoffrey E},
  booktitle={Proceedings of NIPS},
  year={2012}
}

@inproceedings{c:30,
  title={\href{https://proceedings.mlr.press/v202/li23q}{Blip-2: Bootstrapping language-image pre-training with frozen image encoders and large language models}},
  author={Li, Junnan and Li, Dongxu and Savarese, Silvio and Hoi, Steven},
  booktitle={Proceedings of ICML},
  year={2023}
}

@inproceedings{c:31,
  title={\href{https://proceedings.neurips.cc/paper_files/paper/2024/hash/d13a3eae72366e61dfdc7eea82eeb685-Abstract-Conference.html}{Parallelizing linear transformers with the delta rule over sequence length}},
  author={Yang, Songlin and Wang, Bailin and Zhang, Yu and Shen, Yikang and Kim, Yoon},
  booktitle={Proceedings of NIPS},
  year={2024}
}

@inproceedings{c:32,
  title={\href{https://proceedings.neurips.cc/paper_files/paper/2024/hash/d13a3eae72366e61dfdc7eea82eeb685-Abstract-Conference.html}{Gated delta networks: Improving mamba2 with delta rule}},
  author={Yang, Songlin and Kautz, Jan and Hatamizadeh, Ali},
  booktitle={Proceedings of ICLR},
  year={2025}
}

@inproceedings{c:33,
  title={\href{https://openreview.net/forum?id=MSbU3L7V00&noteId=rqP87e9HLl}{Do not think that much for 2+ 3=? on the overthinking of o1-like llms}},
  author={Chen, Xingyu and Xu, Jiahao and Liang, Tian and He, Zhiwei and Pang, Jianhui and Yu, Dian and Song, Linfeng and Liu, Qiuzhi and Zhou, Mengfei and Zhang, Zhuosheng and others},
  booktitle={Proceedings of ICML},
  year={2025}
}

@article{c:34,
  title={\href{https://arxiv.org/abs/2504.15895}{Dynamic Early Exit in Reasoning Models}},
  author={Yang, Chenxu and Si, Qingyi and Duan, Yongjie and Zhu, Zheliang and Zhu, Chenyu and Li, Qiaowei and Lin, Zheng and Cao, Li and Wang, Weiping},
  journal={arXiv:2504.15895},
  year={2025}
}

@article{c:35,
  title={\href{https://ai.meta.com/blog/llama-4-multimodal-intelligence/?utm_source=llama-home-behemoth&utm_medium=llama-referral&utm_campaign=llama-utm&utm_offering=llama-behemoth-preview&utm_product=llama}{The llama 4 herd: The beginning of a new era of natively multimodal ai innovation}},
  author={Meta, AI},
  journal={https://ai. meta. com/blog/llama-4-multimodal-intelligence/, checked on},
  year={2025}
}

@article{c:36,
  title={\href{https://arxiv.org/abs/2505.09388}{Qwen3 technical report}},
  author={Yang, An and Li, Anfeng and Yang, Baosong and Zhang, Beichen and Hui, Binyuan and Zheng, Bo and Yu, Bowen and Gao, Chang and Huang, Chengen and Lv, Chenxu and others},
  journal={arXiv:2505.09388},
  year={2025}
}

@inproceedings{c:37,
  title={\href{https://proceedings.neurips.cc/paper/2017/hash/3f5ee243547dee91fbd053c1c4a845aa-Abstract.html}{Attention is all you need}},
  author={Vaswani, Ashish and Shazeer, Noam and Parmar, Niki and Uszkoreit, Jakob and Jones, Llion and Gomez, Aidan N and Kaiser, {\L}ukasz and Polosukhin, Illia},
  booktitle={Proceedings of NIPS},
  year={2017}
}

@inproceedings{c:38,
  title={\href{https://arxiv.org/abs/2103.13076}{Finetuning pretrained transformers into rnns}},
  author={Kasai, Jungo and Peng, Hao and Zhang, Yizhe and Yogatama, Dani and Ilharco, Gabriel and Pappas, Nikolaos and Mao, Yi and Chen, Weizhu and Smith, Noah A},
  booktitle={Proceedings of EMNLP},
  year={2021}
}

@inproceedings{c:39,
  title={\href{https://arxiv.org/abs/2103.02143}{Random feature attention}},
  author={Peng, Hao and Pappas, Nikolaos and Yogatama, Dani and Schwartz, Roy and Smith, Noah A and Kong, Lingpeng},
  booktitle={Proceedings of ICLR},
  year={2021}
}

@article{c:40,
  title={\href{https://arxiv.org/abs/2307.08621}{Retentive network: A successor to transformer for large language models}},
  author={Sun, Yutao and Dong, Li and Huang, Shaohan and Ma, Shuming and Xia, Yuqing and Xue, Jilong and Wang, Jianyong and Wei, Furu},
  journal={arXiv:2307.08621},
  year={2023}
}

@article{c:41,
  title={\href{https://arxiv.org/abs/2307.08621}{Learning to (learn at test time): Rnns with expressive hidden states}},
  author={Sun, Yu and Li, Xinhao and Dalal, Karan and Xu, Jiarui and Vikram, Arjun and Zhang, Genghan and Dubois, Yann and Chen, Xinlei and Wang, Xiaolong and Koyejo, Sanmi and others},
  journal={arXiv:2407.04620},
  year={2024}
}

@article{c:42,
  title={\href{https://arxiv.org/abs/2507.06203}{A survey on latent reasoning}},
  author={Zhu, Rui-Jie and Peng, Tianhao and Cheng, Tianhao and Qu, Xingwei and Huang, Jinfa and Zhu, Dawei and Wang, Hao and Xue, Kaiwen and Zhang, Xuanliang and Shan, Yong and others},
  journal={arXiv:2507.06203},
  year={2025}
}

@article{c:43,
  title={\href{https://arxiv.org/abs/2501.18585}{Thoughts are all over the place: On the underthinking of o1-like llms}},
  author={Wang, Yue and Liu, Qiuzhi and Xu, Jiahao and Liang, Tian and Chen, Xingyu and He, Zhiwei and Song, Linfeng and Yu, Dian and Li, Juntao and Zhang, Zhuosheng and others},
  journal={arXiv:2501.18585},
  year={2025}
}

@article{c:44,
  title={\href{https://arxiv.org/abs/2506.01939}{Beyond the 80/20 rule: High-entropy minority tokens drive effective reinforcement learning for llm reasoning}},
  author={Wang, Shenzhi and Yu, Le and Gao, Chang and Zheng, Chujie and Liu, Shixuan and Lu, Rui and Dang, Kai and Chen, Xionghui and Yang, Jianxin and Zhang, Zhenru and others},
  journal={arXiv:2506.01939},
  year={2025}
}

@inproceedings{c:45,
  title={\href{https://arxiv.org/abs/2103.13076}{Lora: Low-rank adaptation of large language models}},
  author={Hu, Edward J and Shen, Yelong and Wallis, Phillip and Allen-Zhu, Zeyuan and Li, Yuanzhi and Wang, Shean and Wang, Lu and Chen, Weizhu and others},
  booktitle={Proceedings of ICLR},
  year={2022}
}

@inproceedings{c:46,
  title={\href{https://arxiv.org/abs/2312.06635}{Gated linear attention transformers with hardware-efficient training}},
  author={Yang, Songlin and Wang, Bailin and Shen, Yikang and Panda, Rameswar and Kim, Yoon},
  booktitle={Proceedings of ICML},
  year={2024}
}

@inproceedings{c:47,
  title={\href{https://arxiv.org/abs/2404.07904}{Hgrn2: Gated linear rnns with state expansion}},
  author={Qin, Zhen and Yang, Songlin and Sun, Weixuan and Shen, Xuyang and Li, Dong and Sun, Weigao and Zhong, Yiran},
  booktitle={Proceedings of COLM},
  year={2024}
}

@inproceedings{c:48,
  title={\href{https://arxiv.org/abs/2503.01496}{Liger: Linearizing Large Language Models to Gated Recurrent Structures}},
  author={Lan, Disen and Sun, Weigao and Hu, Jiaxi and Du, Jusen and Cheng, Yu},
  booktitle={Proceedings of ICML},
  year={2025}
}

@inproceedings{c:49,
  title={\href{https://arxiv.org/abs/2407.14207}{Longhorn: State space models are amortized online learners}},
  author={Liu, Bo and Wang, Rui and Wu, Lemeng and Feng, Yihao and Stone, Peter and Liu, Qiang},
  booktitle={Proceedings of ICLR},
  year={2025}
}

@article{c:50,
  title={\href{https://arxiv.org/abs/2503.06692}{Inftythink: Breaking the length limits of long-context reasoning in large language models}},
  author={Yan, Yuchen and Shen, Yongliang and Liu, Yang and Jiang, Jin and Zhang, Mengdi and Shao, Jian and Zhuang, Yueting},
  journal={arXiv:2503.06692},
  year={2025}
}

@inproceedings{c:51,
  title={\href{https://openreview.net/forum?id=RkRrPp7GKO}{H2o: Heavy-hitter oracle for efficient generative inference of large language models}},
  author={Zhang, Zhenyu and Sheng, Ying and Zhou, Tianyi and Chen, Tianlong and Zheng, Lianmin and Cai, Ruisi and Song, Zhao and Tian, Yuandong and R{\'e}, Christopher and Barrett, Clark and others},
  booktitle={Proceedings of NIPS},
  year={2023}
}

@inproceedings{c:52,
  title={\href{https://arxiv.org/abs/2412.12094}{Sepllm: Accelerate large language models by compressing one segment into one separator}},
  author={Chen, Guoxuan and Shi, Han and Li, Jiawei and Gao, Yihang and Ren, Xiaozhe and Chen, Yimeng and Jiang, Xin and Li, Zhenguo and Liu, Weiyang and Huang, Chao},
  booktitle={Proceedings of ICML},
  year={2025}
}

@article{c:53,
  title={\href{https://arxiv.org/abs/2110.14168}{Training verifiers to solve math word problems}},
  author={Cobbe, Karl and Kosaraju, Vineet and Bavarian, Mohammad and Chen, Mark and Jun, Heewoo and Kaiser, Lukasz and Plappert, Matthias and Tworek, Jerry and Hilton, Jacob and Nakano, Reiichiro and others},
  journal={arXiv:2110.14168},
  year={2021}
}

@inproceedings{c:54,
  title={\href{https://arxiv.org/abs/2103.03874}{Measuring mathematical problem solving with the math dataset}},
  author={Hendrycks, Dan and Burns, Collin and Kadavath, Saurav and Arora, Akul and Basart, Steven and Tang, Eric and Song, Dawn and Steinhardt, Jacob},
  booktitle={Proceedings of NIPS},
  year={2021}
}

@inproceedings{c:55,
  title={\href{https://openreview.net/forum?id=Ti67584b98}{Gpqa: A graduate-level google-proof q\&a benchmark}},
  author={Rein, David and Hou, Betty Li and Stickland, Asa Cooper and Petty, Jackson and Pang, Richard Yuanzhe and Dirani, Julien and Michael, Julian and Bowman, Samuel R},
  booktitle={Proceedings of COLM},
  year={2024}
}

@article{c:56,
  title={\href{https://arxiv.org/abs/2107.03374}{Evaluating large language models trained on code}},
  author={Chen, Mark and Tworek, Jerry and Jun, Heewoo and Yuan, Qiming and Pinto, Henrique Ponde De Oliveira and Kaplan, Jared and Edwards, Harri and Burda, Yuri and Joseph, Nicholas and Brockman, Greg and others},
  journal={arXiv:2107.03374},
  year={2021}
}

@online{aime2024,
  author       = {{AIME}},
  title        = {American invitational mathematics examination (aime) aime 2024-i \& ii},
  year         = {2024},
  url          = {https://huggingface.co/datasets/Maxwell-Jia/AIME_2024}
}

@online{aime2025,
  author       = {{AIME}},
  title        = {American invitational mathematics examination (aime) 2025-i \& ii},
  year         = {2025},
  url          = {https://huggingface.co/datasets/opencompass/AIME2025}
}

@online{amc2023,
  author       = {{AMC}},
  title        = {American mathematics competitions},
  year         = {2023},
  url          = {https://artofproblemsolving.com/wiki/index.php/AMC_Problems_and_Solutions}
}

@article{c:57,
  title={\href{https://arxiv.org/abs/2307.08691}{Flashattention-2: Faster attention with better parallelism and work partitioning}},
  author={Dao, Tri},
  journal={arXiv:2307.08691},
  year={2023}
}

@article{c:58,
  title={\href{https://arxiv.org/abs/1905.12322}{A study of BFLOAT16 for deep learning training}},
  author={Kalamkar, Dhiraj and Mudigere, Dheevatsa and Mellempudi, Naveen and Das, Dipankar and Banerjee, Kunal and Avancha, Sasikanth and Vooturi, Dharma Teja and Jammalamadaka, Nataraj and Huang, Jianyu and Yuen, Hector and others},
  journal={arXiv:1905.12322},
  year={2019}
}

@article{c:59,
  title={\href{https://arxiv.org/abs/2505.16315}{Incentivizing Dual Process Thinking for Efficient Large Language Model Reasoning}},
  author={Cheng, Xiaoxue and Li, Junyi and Zhang, Zhenduo and Tang, Xinyu and Zhao, Wayne Xin and Kong, Xinyu and Zhang, Zhiqiang},
  journal={arXiv:2505.16315},
  year={2025}
}

@article{c:60,
  title={\href{https://arxiv.org/abs/2503.20641}{Unlocking efficient long-to-short llm reasoning with model merging}},
  author={Wu, Han and Yao, Yuxuan and Liu, Shuqi and Liu, Zehua and Fu, Xiaojin and Han, Xiongwei and Li, Xing and Zhen, Hui-Ling and Zhong, Tao and Yuan, Mingxuan},
  journal={arXiv preprint arXiv:2503.20641},
  year={2025}
}

@inproceedings{c:61,
  title={\href{https://arxiv.org/abs/2503.04697}{L1: Controlling how long a reasoning model thinks with reinforcement learning}},
  author={Aggarwal, Pranjal and Welleck, Sean},
  booktitle={Proceedings of COLM},
  year={2025}
}

@article{c:62,
  title={\href{https://arxiv.org/abs/2503.15952}{Adaptive group policy optimization: Towards stable training and token-efficient reasoning}},
  author={Li, Chen and Liu, Nazhou and Yang, Kai},
  journal={arXiv:2503.15952},
  year={2025}
}

@article{c:63,
  title={\href{https://arxiv.org/abs/2504.05185}{Concise reasoning via reinforcement learning}},
  author={Fatemi, Mehdi and Rafiee, Banafsheh and Tang, Mingjie and Talamadupula, Kartik},
  journal={arXiv:2504.05185},
  year={2025}
}

@article{c:64,
  title={\href{https://arxiv.org/abs/2505.11274}{Selfbudgeter: Adaptive token allocation for efficient llm reasoning}},
  author={Li, Zheng and Dong, Qingxiu and Ma, Jingyuan and Zhang, Di and Jia, Kai and Sui, Zhifang},
  journal={arXiv:2505.11274},
  year={2025}
}

@article{c:65,
  title={\href{https://arxiv.org/abs/2505.16122}{Plan and Budget: Effective and Efficient Test-Time Scaling on Large Language Model Reasoning}},
  author={Lin, Junhong and Zeng, Xinyue and Zhu, Jie and Wang, Song and Shun, Julian and Wu, Jun and Zhou, Dawei},
  journal={arXiv:2505.16122},
  year={2025}
}

@article{c:66,
  title={\href{https://arxiv.org/abs/2505.17827}{Not All Tokens Are What You Need In Thinking}},
  author={Yuan, Hang and Yu, Bin and Li, Haotian and Yang, Shijun and Wang, Christina Dan and Yu, Zhou and Xu, Xueyin and Qi, Weizhen and Chen, Kai},
  journal={arXiv:2505.17827},
  year={2025}
}

@article{c:67,
  title={\href{https://arxiv.org/abs/2505.18086}{Stable Reinforcement Learning for Efficient Reasoning}},
  author={Dai, Muzhi and Liu, Shixuan and Si, Qingyi},
  journal={arXiv:2505.18086},
  year={2025}
}

@article{c:68,
  title={\href{https://arxiv.org/abs/2506.05256}{Just Enough Thinking: Efficient Reasoning with Adaptive Length Penalties Reinforcement Learning}},
  author={Xiang, Violet and Blagden, Chase and Rafailov, Rafael and Lile, Nathan and Truong, Sang and Finn, Chelsea and Haber, Nick},
  journal={arXiv preprint arXiv:2506.05256},
  year={2025}
}

@article{c:69,
  title={\href{https://arxiv.org/abs/2505.13438}{Optimizing anytime reasoning via budget relative policy optimization}},
  author={Qi, Penghui and Liu, Zichen and Pang, Tianyu and Du, Chao and Lee, Wee Sun and Lin, Min},
  journal={arXiv:2505.13438},
  year={2025}
}

@article{c:70,
  title={\href{https://arxiv.org/abs/2505.07961}{Making small language models efficient reasoners: Intervention, supervision, reinforcement}},
  author={Zhang, Xuechen and Huang, Zijian and Ni, Chenshun and Xiong, Ziyang and Chen, Jiasi and Oymak, Samet},
  journal={arXiv:2505.07961},
  year={2025}
}

@article{c:71,
  title={\href{https://arxiv.org/abs/2505.18237}{Think or Not? Exploring Thinking Efficiency in Large Reasoning Models via an Information-Theoretic Lens}},
  author={Yong, Xixian and Zhou, Xiao and Zhang, Yingying and Li, Jinlin and Zheng, Yefeng and Wu, Xian},
  journal={arXiv:2505.18237},
  year={2025}
}

@article{c:72,
  title={\href{https://arxiv.org/abs/2505.14604}{Let LLMs Break Free from Overthinking via Self-Braking Tuning}},
  author={Zhao, Haoran and Yan, Yuchen and Shen, Yongliang and Xu, Haolei and Zhang, Wenqi and Song, Kaitao and Shao, Jian and Lu, Weiming and Xiao, Jun and Zhuang, Yueting},
  journal={arXiv:2505.14604},
  year={2025}
}

@article{c:73,
  title={\href{https://arxiv.org/abs/2505.17941}{VeriThinker: Learning to Verify Makes Reasoning Model Efficient}},
  author={Chen, Zigeng and Ma, Xinyin and Fang, Gongfan and Yu, Ruonan and Wang, Xinchao},
  journal={arXiv:2505.17941},
  year={2025}
}

@article{c:74,
  title={\href{https://arxiv.org/abs/2505.13438}{Optimizing anytime reasoning via budget relative policy optimization}},
  author={Qi, Penghui and Liu, Zichen and Pang, Tianyu and Du, Chao and Lee, Wee Sun and Lin, Min},
  journal={arXiv:2505.13438},
  year={2025}
}

@article{c:75,
  title={\href{https://arxiv.org/abs/2505.07686}{S-GRPO: Early Exit via Reinforcement Learning in Reasoning Models}},
  author={Dai, Muzhi and Yang, Chenxu and Si, Qingyi},
  journal={arXiv:2505.07686},
  year={2025}
}

@article{c:76,
  title={\href{https://arxiv.org/abs/2508.02511}{Test-time Prompt Intervention}},
  author={Yang, Chenxu and Si, Qingyi and Dai, Mz and Yao, Dingyu and Zheng, Mingyu and Chen, Minghui and Lin, Zheng and Wang, Weiping},
  journal={arXiv preprint arXiv:2508.02511},
  year={2025}
}

@article{c:77,
  title={\href{https://arxiv.org/abs/2511.00405}{UME-R1: Exploring Reasoning-Driven Generative Multimodal Embeddings}},
  author={Lan, Zhibin and Niu, Liqiang and Meng, Fandong and Zhou, Jie and Su, Jinsong},
  journal={arXiv:2511.00405},
  year={2025}
}

@inproceedings{c:78,
  title={\href{https://ojs.aaai.org/index.php/AAAI/article/view/34719}{LiteSearch: Efficient Tree Search with Dynamic Exploration Budget for Math Reasoning}},
  author={Wang, Ante and Song, Linfeng and Tian, Ye and Peng, Baolin and Yu, Dian and Mi, Haitao and Su, Jinsong and Yu, Dong},
  booktitle={Proceedings of AAAI},
  year={2025}
}

@article{c:79,
  title={\href{https://arxiv.org/abs/2403.09849}{Self-consistency boosts calibration for math reasoning}},
  author={Wang, Ante and Song, Linfeng and Tian, Ye and Peng, Baolin and Jin, Lifeng and Mi, Haitao and Su, Jinsong and Yu, Dong},
  journal={arXiv preprint arXiv:2403.09849},
  year={2024}
}

@inproceedings{c:80,
  title={\href{https://aclanthology.org/P18-1166/}{Accelerating neural transformer via an average attention network}},
  author={Zhang, Biao and Xiong, Deyi and Su, Jinsong},
  booktitle={Proceedings of ACL},
  year={2018}
}

@article{c:81,
  title={\href{https://www.jair.org/index.php/jair/article/view/13896}{Aan+: Generalized average attention network for accelerating neural transformer}},
  author={Zhang, Biao and Xiong, Deyi and Ge, Yubin and Yao, Junfeng and Yue, Hao and Su, Jinsong},
  journal={Journal of Artificial Intelligence Research},
  year={2022}
}
\bibliographystyle{iclr2026_conference}

\appendix
\section{Further Analysis}
\label{appendix:A}
\begin{figure}[ht]
\centering
\includegraphics[width=1.0 \textwidth]{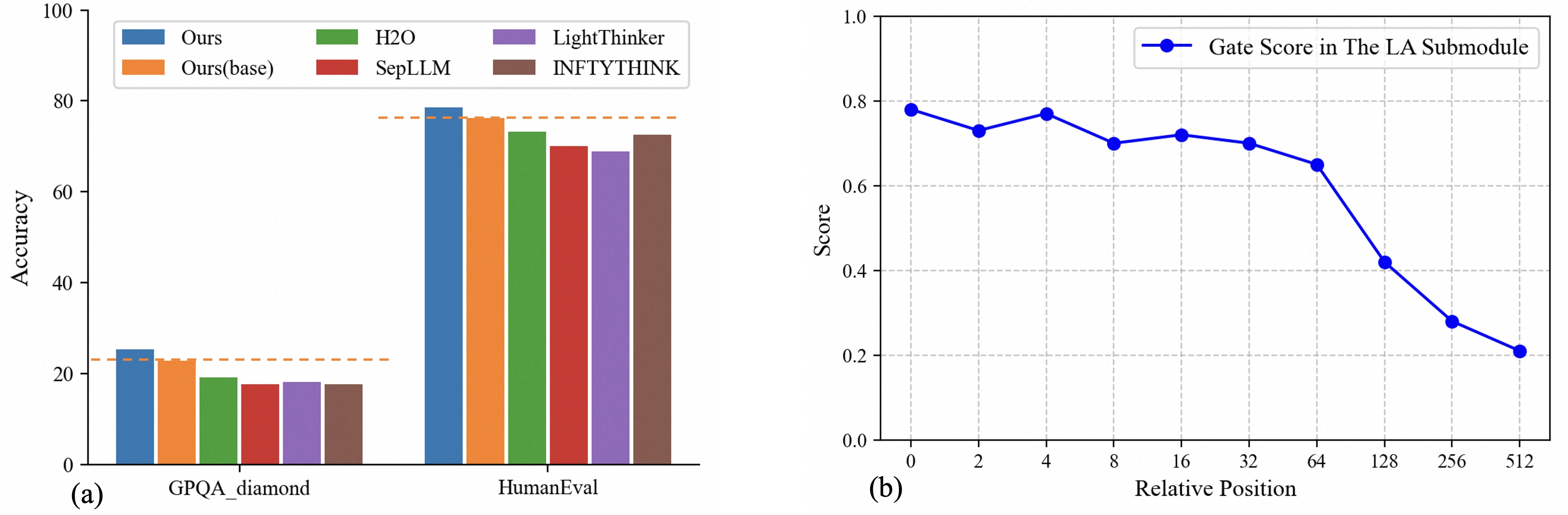}
\caption{(a) shows the performance of our model and the baselines on the GPQA Diamond and HumanEval benchmarks.
(b) illustrates the average gating scores in the LA submodule for tokens at different positions within each reasoning step.
These experiments are conducted on Qwen2.5-7B.
}
\label{fig5}
\end{figure}

\textbf{Analysis of Domain Generalization.}
As stated in Section~\ref{sec:5.1}, our training set contains only mathematical reasoning data.
To examine the domain generalization of our framework, we further test our model on the scientific reasoning benchmark GPQA\_Diamond and the code reasoning benchmark HumanEval.
As shown in Figure~\ref{fig5}(a), our model consistently outperforms all baselines on these two benchmarks.
% These results demonstrate that our model state can effectively store domain-agnostic important historical reasoning information, which ensures the accuracy of subsequent reasoning.
These results reveal several advantages of our framework: (1) our training strategy effectively prevents the negative impact of target-domain training on the base model’s performance on other domains; (2) our framework can effectively transfer the performance and efficiency benefits in the target domain to other domains.

\textbf{Analysis of Gating Scores.}
Here, we further investigate how much historical reasoning information is required by tokens at different positions within each reasoning step.
To achieve this, we calculate the average gating scores of these tokens in the LA submodule.
As illustrated in Figure~\ref{fig5}(b), tokens occurring earlier in a reasoning step generally require more historical reasoning information from the model state than those occurring later.
The main reason is that  later tokens can directly obtain more information from earlier tokens via the SA sublayer, reducing their reliance on the model state.

\section{Implementation Details}
\label{appendix:B}
We train both our model and the baselines for 2 epochs on the training set, with a batch size of 32.
The learning rate is set to $2e{-}4$, with a warmup ratio of 0.03.
The learning rate follows a cosine decay schedule to reach zero.
We set the learnable LoRA parameters for both our model and the baselines to approximately 66.5M on Qwen2.5‑1.5B and 147.3M on Qwen2.5‑7B, respectively.
To control training cost, we limit the maximum length of long CoT samples in the training set to 16,384 tokens.
Meanwhile, we train our model and the baselines on 8 NVIDIA A100 GPUs, each with 80GB of memory.
Further implementation details are provided below:
\begin{itemize}
  \item \textbf{First}, to minimize conflicts in thinking types, we employ a K-means clustering algorithm with 128 clusters to annotate the thinking type of each step. A larger number of clusters can effectively reduce the likelihood of assigning reasoning steps with different thinking patterns to the same cluster. Notably, given the large number of reasoning steps (1.4M), we utilize MiniBatch K-Means clustering to improve clustering efficiency. Meanwhile, We use cosine similarity as the distance metric, resulting in a balanced distribution of reasoning steps across clusters (max: 16.1k; min: 4.3k). 
  \item \textbf{Second}, we adopt the LoRA strategy to implement our linear attention module, where the rank $r$ is used to control the number of learnable parameters. Specifically, we set the rank $r$ of $W_q$, $W_k$, $W_v$ to 512, and the rank $r$ of $W_g$ to 16.
  \item \textbf{Third}, in linear attention, the dimension $d$ of the model state matrix $S_t {=} S_{t-1} {+} k_{t}^T v_t \in R^{d \times d}$ corresponds to the dimensionality of the key and value vectors in LLMs, which is determined by the model architecture itself.
  \item \textbf{Third} as only the newly introduced parameters (the linear attention module and special tokens) are trained, the training cost of our model is reduced to \textbf{47.7\%} of that of the base model. Concretely, with 8 NVIDIA A100 GPUs and the Qwen‑2.5‑1.5B model configuration, one training epoch requires 8.6 hours for the base model and 4.1 hours for our model.
\end{itemize}

\begin{figure}[t]
\centering
\includegraphics[width=1.0 \textwidth]{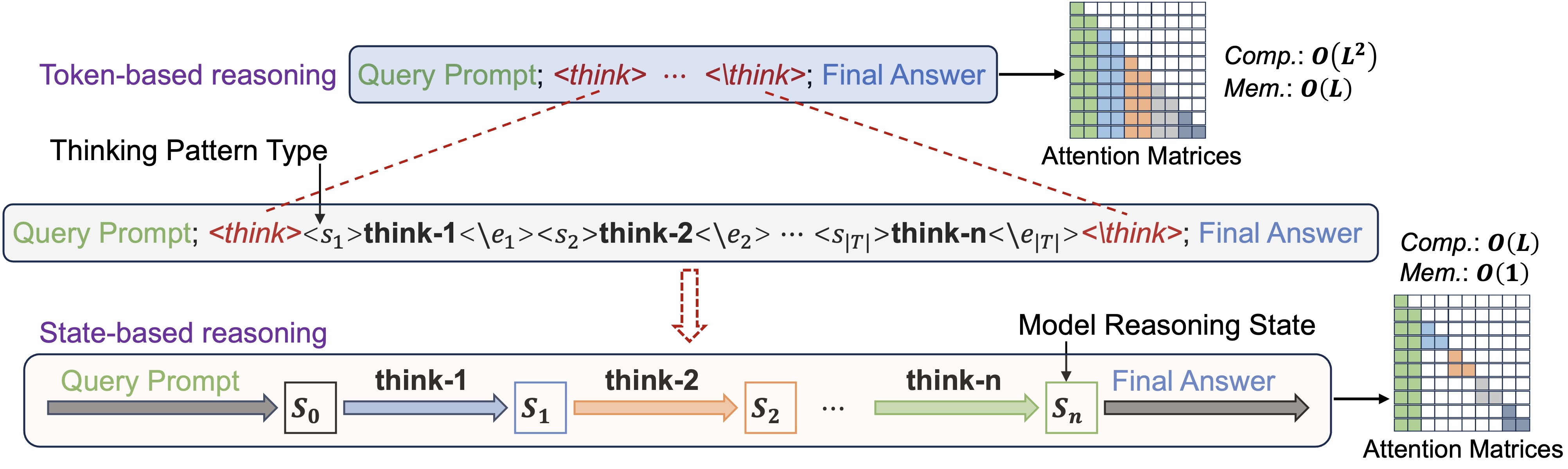}
\caption{A comparison of traditional {\em token‑based reasoning} with our {\em state‑based reasoning}. $\textbf{think}_{t}$ denotes one reasoning step. 
\textbf{Comp.} and \textbf{Mem.} represent the computational and memory complexity, respectively, with $L$ indicating the context length.
In state-based reasoning, LLMs can efficiently generate $\textbf{think}_{t}$ only based on the query prompt and the reasoning state $\boldsymbol{S}_{t{-}1}$.
}
\label{fig9}
\end{figure}

\section{Further analysis of our framework}
\label{appendix:F}
In Figure~\ref{fig9}, we present a detailed comparison of the reasoning processes between our model (state-based reasoning) and naive LLMs (token-based reasoning). 
In the case of a naive LLM, we provide it with a query, after which it autoregressively generates the full CoT, where each token attends to all previously generated tokens.
Meanwhile, the key and value vectors of all generated tokens must be stored in the KV‑cache.
Therefore, under this setting, the attention computational complexity scales quadratically with the CoT length $L$ (i.e., $O(L^2)$), while the memory complexity scales linearly (i.e., $O(L)$).

In our framework, LLMs generate the entire CoT step by step, in units of reasoning steps: $\textbf{think}_{1} \to \cdots \to \textbf{think}_{n}$.
Meanwhile, we adopt a linear attention mechanism to estimate the state $\boldsymbol{S}$ of LLMs during the reasoning process. 
As showed in Figure~\ref{fig9} (bottom), as LLM reasoning progresses ($\textbf{think}_{1} \to \cdots \to \textbf{think}_{n}$), its state undergoes continuous transitions ($\boldsymbol{S}_{0} \to \cdots \to \boldsymbol{S}_{n}$).
Therefore, in our framework, the LLM’s reasoning process is modeled as a state‑transition process.
Importantly, the state of our model performs two key roles:
\begin{itemize}
  \item \textbf{First}, the state records/stores the historical reasoning information of the reasoning steps that have been completed. Hence, our model can generate the next reasoning step ($\textbf{think}_{t}$)  based solely on the current state ($\boldsymbol{S}_{t-1}$) and the query, without depending on any previously completed steps ($\textbf{think}_{<t}$).
  In this way, we reduce the computational complexity of attention in LLMs to linear, $O(L)$.
  Since previously finished reasoning steps are not needed for subsequent reasoning, the key and value vectors of the tokens in these steps are removed from the KV‑cache.
  As a result, the storage complexity of attention in LLMs is reduced to $O(1)$.
  \item \textbf{Second}, the model state can be interpreted as a quantitative representation of the LLM’s knowledge. Consequently, in our state-based reasoning strategy, we correct the reasoning direction of noisy steps to prevent them from contaminating the LLMs’ knowledge (state), thus improving the reasoning performance. Moreover, we are conducting research into using model states to unsupervisedly estimate process rewards for reasoning steps, enhancing the efficiency and effectiveness of LLMs in RL training.
\end{itemize}

\begin{figure}[t]
\centering
\includegraphics[width=1.0 \textwidth]{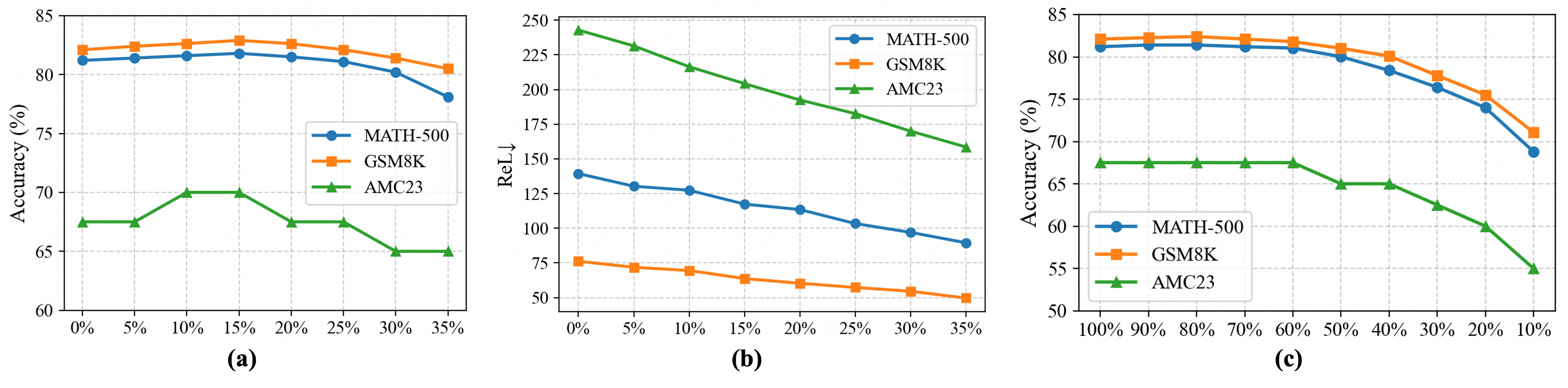}
\caption{(a) and (b) show the performance and reasoning latency of our model versus the fraction of reasoning steps removed. (c) illustrates our model’s performance after removing varying proportions of training samples.
These experiments are conducted on Qwen2.5-1.5B.
}
\label{fig9-1}
\end{figure}

Here, we also conducted an analytical experiment to further validate the potential of our model state.
Specifically, for each sample in the training set, the process reward of every reasoning step is estimated by computing the cosine of the angle between the global reasoning direction $\bar{\nabla}$ and the step’s reasoning direction ${\nabla}_t$ (i.e., $\cos(\bar{\nabla},{\nabla}_t)$).
Next, we remove reasoning steps with low process rewards from each sample to shorten the CoT length.
Finally, we present the experimental results in Figure~\ref{fig9-1}.
We observe that removing less than 20\% of the steps resulted in a performance improvement for our model.
This indicates that such simple process rewards can reflect the quality of reasoning steps.
Moreover, when more than 25\% of the steps are removed, our model’s performance begins to decline.
This may be attributed to the removal of critical steps in the CoT, which limits the model’s reasoning capacity.
Meanwhile, shortening the CoT length further improves the reasoning efficiency of our model.

In addition, we further use our model state to estimate training‑sample quality, thereby improving the data efficiency of our method.
Specifically, we first calculate the global process reward of a training sample as the average of the process rewards from its reasoning steps.
The global process reward reflects the extent to which a sample contains noisy reasoning steps.
Then, we discard training samples with low global process rewards and retrain our model. 
As shown in Figure~\ref{fig9-1}(c), the performance of our model does not degrade when 60\% of high‑quality training samples are retained.
Meanwhile, removing  60\% of the training samples (i.e., retaining 40
\%) results in only a slight performance reduction.

\section{Generalization Analysis of the CoT Segmentation Method}
\label{appendix:H}
Indeed, our long CoT segmentation method is also a general framework.
The framework includes two primary operations:
\begin{itemize}
  \item \textbf{Segmentation,} which aims to segment a long CoT into a sequence of reasoning steps.
  \item \textbf{Annotation,} focused on identifying the thinking type underlying each reasoning step.
\end{itemize}
Meanwhile, it adheres to two core principles:
 \begin{itemize}
  \item \textbf{First,} segmented reasoning steps should maintain low linguistic dependency, mitigating information loss caused by KV-cache cleaning.
  \item \textbf{Second,} conflicts among thinking types should be minimized to avoid detrimental effects from type incompatibilities.
\end{itemize}

The primary objective of the framework is to enhance the controllability of LLM reasoning processes.
In this setting, we can utilize special tokens, associated with the thinking types, to monitor and control the reasoning processes of LLMs.

With sufficient resources, CoT segmentation and thinking-type annotation can be conducted manually, yielding superior performance. An alternative viable approach is to employ powerful LLMs, such as GPT‑4o, to carry out segmentation and annotation.

Given resource limitations, we adopt a more economical strategy: (1) segmentation is performed using high-entropy transition tokens appearing at the beginning of sentences, ensuring that each reasoning step remains semantically independent; and (2) we conduct thinking‑type annotation via a clustering algorithm (K-means), while configuring it with a large number of clusters (128) to minimize inter‑type conflicts. Nevertheless, our model achieves outstanding performance in the math domain while generalizing effectively to both scientific and code domains.

Furthermore, the details of the clustering algorithm we used are as follows:

 \begin{itemize}
  \item Given the large number of reasoning steps (1.4M), we utilize MiniBatch K-Means clustering to improve clustering efficiency..
  \item We used a larger number of clusters (128) to effectively reduce the probability of grouping reasoning steps with distinct thinking patterns into the same cluster.
  \item We use cosine similarity as the distance metric, resulting in a balanced distribution of reasoning steps across clusters (max: 16.1k; min: 4.3k). This effectively reduces the risk of LLMs "over-fitting" or "under-fitting" to thinking patterns.
\end{itemize}

\begin{table*}[t]
% \small
\setlength{\abovecaptionskip}{6pt}
\centering
\setlength{\tabcolsep}{6pt}
\resizebox{\textwidth}{!}{
\begin{tabular}{lccccccccc}
\toprule[1.2pt]
{\textbf{Method}}
 & \multicolumn{3}{c}{\textbf{MATH-500}}
 & \multicolumn{3}{c}{\textbf{AMC23}}
 & \multicolumn{3}{c}{\textbf{AIME24}} \\
\midrule[0.9pt]
\multicolumn{10}{c}{\textit{Qwen2.5-1.5B Series}} \\
\midrule[0.9pt]
\textit{Greedy Decoding}  & Acc$\uparrow$ & Tok & RC\_rate$\downarrow$  
 & Acc$\uparrow$ & Tok & RC\_rate$\downarrow$  
 & Acc$\uparrow$ & Tok & RC\_rate$\downarrow$ \\
\midrule[0.9pt]
\rowcolor{lightgray}Ours (Base)  
 & 78.8 & 3958 & 0.31	& 62.5 & 6392 & 0.40	& 20.0 & 13765 & 0.50 \\
\rowcolor{lightblue}Ours        
 & 81.2 & 3812 & 0.18	& 67.5 & 6460 & 0.23	& 26.7 & 13610 & 0.31 \\
\midrule[0.9pt]
\textit{Sampling Decoding}  & pass@1$\uparrow$ & Tok & RC\_rate$\downarrow$  
 &  pass@1$\uparrow$ & Tok & RC\_rate$\downarrow$  
 &  pass@1$\uparrow$ & Tok & RC\_rate$\downarrow$ \\
\midrule[0.9pt]
\rowcolor{lightgray}Ours (Base)  
 &80.3 &3780 &0.19	&63.6 &6607 &0.28	&22.5 &13880 &0.26 \\
\rowcolor{lightblue}Ours        
  &83.5  &3652  &0.06	  &69.4  &6533  &0.08	 &29.1   &13809  &0.09 \\
\bottomrule[1.2pt]
\end{tabular}
}
\caption{The results of our model and baselines under different decoding strategies. RC\_rate denotes the proportion of the question for which the model failed to reach the final answer due to repeated generation out of all incorrect questions.}
\label{table:1-2}
\end{table*}

\section{Further analysis of model performance}
\label{appendix:I}

\textbf{Analysis of Decoding Strategies.} As observed in DeepSeek‑R1 \cite{c:7}, a temperature‑based sampling strategy can effectively reduce the proportion of repeated generation, thus improving the reasoning performance of LLMs.
Here, we adopted the DeepSeek‑R1 sampling strategy (temperature=0.6, top‑k=0.95, run 32 times) and presented pass@1 in Table~\ref{table:1-2}.
We observed performance improvements in both our model and the base model. The primary reason is that this sampling method significantly reduces the proportion of repeated generation (i.e., \textit{RC rate}).
To avoid overestimating the performance of our model and affecting the reproducibility of our work, we adopted a more deterministic greedy decoding method in the main experiment.

\begin{table*}[t]
\setlength{\abovecaptionskip}{6pt}
\centering
\setlength{\tabcolsep}{1.8pt}
\resizebox{\textwidth}{!}{
\begin{tabular}{lccccccccccccccccc}
\toprule[1.2pt]
\multirow{2}{*}{\textbf{Method}}
 & \multicolumn{3}{c}{\textbf{GSM8K}} 
 & \multicolumn{3}{c}{\textbf{MATH-500}}
 & \multicolumn{3}{c}{\textbf{AMC23}}
 & \multicolumn{3}{c}{\textbf{AIME24}}
 & \multicolumn{3}{c}{\textbf{AIME25}}
 & \multicolumn{2}{c}{\textbf{AVG.}} \\
\cmidrule(lr){2-4} \cmidrule(lr){5-7} \cmidrule(lr){8-10} \cmidrule(lr){11-13} \cmidrule(lr){14-16} \cmidrule(lr){17-18}
 & Acc$\uparrow$ & Tok & ReL$\downarrow$ 
 & Acc$\uparrow$ & Tok & ReL$\downarrow$  
 & Acc$\uparrow$ & Tok & ReL$\downarrow$  
 & Acc$\uparrow$ & Tok & ReL$\downarrow$  
 & Acc$\uparrow$ & Tok & ReL$\downarrow$
 & Acc$\uparrow$ & ReL$\downarrow$ \\
\midrule[0.9pt]
\multicolumn{18}{c}{\textit{Qwen2.5-7B Series} ($\beta=0.4$, $\alpha=0.3$)} \\
\midrule[0.9pt]
\rowcolor{lightgray}Ours (Base)          &89.4 &2649 &96.4	&87.4 &4253 &161.6	&82.5 &6704 &242.9	&40.0 &12901 &522.1	&30.0 &13204 &548.1	&65.9 &314.2   \\
\rowcolor{lightblue}
% \rowcolor{blue!16}
\textbf{Ours}    &91.6  &2580  &86.4   &90.4  &4244  &142.2  &87.5  &6602  &211.7  &43.3  &12730  &432.0  &36.7 &13310  &442.2     &69.9 &262.9  \\
\bottomrule[1.3pt]
\end{tabular}
}
\caption{The results of our model and baselines on mathematical reasoning benchmarks.}
\label{table:1-4}
\end{table*}

\textbf{Analysis of Performance-Improvement Magnitude.}
Notably, the performance gains achieved by our model is non-trivial:
 \begin{itemize}
\item \textbf{First}, in the Qwen‑2.5‑1.5B scenario, our model achieves an average improvement of 4.7 points over the base model.
\item \textbf{Second}, although hyperparameters were selected for Qwen‑2.5‑1.5B, our model delivers an average gain of 2.4 points over the base model in the Qwen‑2.5‑7B configuration. Naturally, conducting more targeted hyperparameter tuning will further yield greater performance gains (4.0 points) for our model on Qwen‑2.5‑7B (See Table~\ref{table:1-4}).
\item \textbf{Third}, the main goal of our work is to improve the reasoning efficiency of LLMs, while performance gains are an unexpected surprise.
\end{itemize}

\section{Case Study}
\label{appendix:G}

\begin{figure}[ht]
\setlength{\abovecaptionskip}{4pt}
\setlength{\belowcaptionskip}{-8pt}
\centering
\includegraphics[width=1.0 \textwidth]{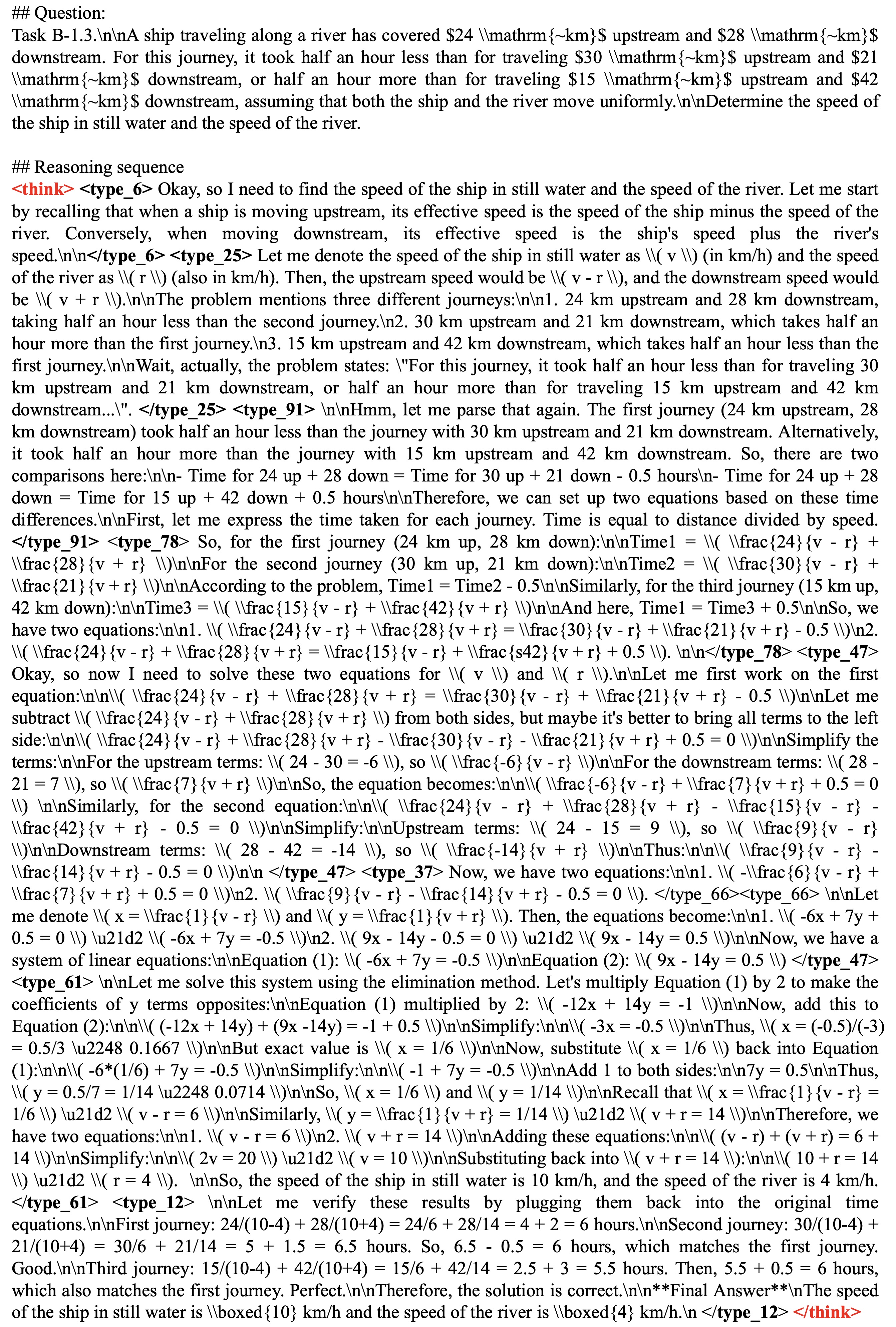}
\caption{\textbf{Case-1} $<$type\_i$>$ and $<$type\_i$>$ denote the start and end tokens, respectively, for reasoning steps corresponding to the $i$-th thinking type.
}
\label{fig10}
\end{figure}

\begin{figure}[ht]
\centering
\includegraphics[width=1.0 \textwidth]{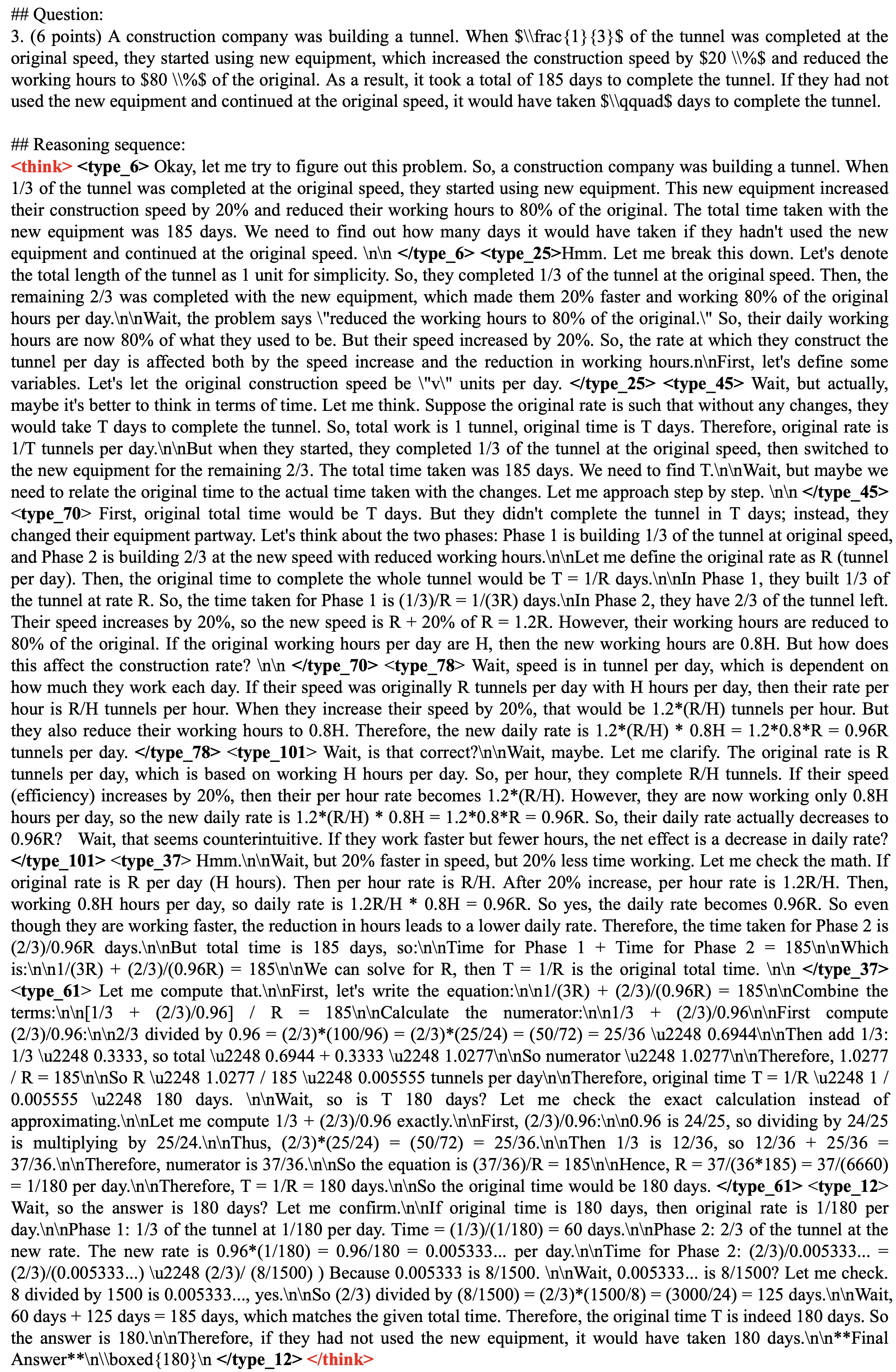}
\caption{\textbf{Case-2:} $<$type\_i$>$ and $<$type\_i$>$ denote the start and end tokens, respectively, for reasoning steps corresponding to the $i$-th thinking type.}
\label{fig11}
\end{figure}

In Figure~\ref{fig10} and \ref{fig11}, we present two case studies to illustrate our model’s reasoning process.
We note that both cases involve 8 and 9 distinct types of thinking, respectively.
Moreover, the two cases also share several thinking types:
\begin{itemize}
  \item \textbf{$<$type\_6$>$:} problem understanding.
  \item \textbf{$<$type\_25$>$:} problem interpretation.
  \item \textbf{$<$type\_61$>$:} algebraic solution.
  \item \textbf{$<$type\_12$>$:} solution verification.
    \item $\cdots$
\end{itemize}
We obtain the interpretations for these thinking types using GPT‑4o.
Moreover, we observe that many reasoning types are general‑purpose and not restricted to the mathematical domain, such as \textbf{$<$type\_6$>$}, \textbf{$<$type\_25$>$}, and \textbf{$<$type\_12$>$}.
This also indirectly illustrates the domain‑generalization capability of our framework.

\end{document}